\pdfoutput=1

\documentclass[11pt]{article}

\usepackage[final]{acl}

\usepackage{times}
\usepackage{latexsym}

\usepackage[T1]{fontenc}

\usepackage[utf8]{inputenc}

\usepackage{microtype}

\usepackage{inconsolata}

\usepackage{graphicx}

\usepackage{amsmath,amsfonts,bm}

\def\eqref#1{equation~\ref{#1}}

\def\1{\bm{1}}

\DeclareMathAlphabet{\mathsfit}{\encodingdefault}{\sfdefault}{m}{sl}
\SetMathAlphabet{\mathsfit}{bold}{\encodingdefault}{\sfdefault}{bx}{n}

\usepackage{hyperref}
\usepackage{url}

\usepackage{amsmath}
\usepackage{amssymb}
\usepackage{url}
\usepackage{booktabs} 
\usepackage{graphicx}
\usepackage{epstopdf}
\usepackage{verbatim}
\usepackage{xcolor}         
\usepackage{bm}
\usepackage{array}
\usepackage{multirow}
\usepackage{makecell}
\usepackage{xcolor}
\usepackage{xspace}
\usepackage{framed}
\usepackage{changepage}
\usepackage{cleveref}
\usepackage{hyperref}
\usepackage{enumitem}
\usepackage{multirow}
\usepackage{booktabs}
\usepackage{subcaption}   
\usepackage{tabularx}
\usepackage{float}  
\usepackage{placeins}  

\newcommand{\RN}[1]{%
  \textup{\uppercase\expandafter{\romannumeral#1}}%
}

\newcommand{\data}{\textsc{CBT-Bench}\xspace}
\newcommand{\cbtqa}{\textsc{CBT-QA}\xspace}
\newcommand{\cbtcd}{\textsc{CBT-CD}\xspace}
\newcommand{\cbtmc}{\textsc{CBT-PC}\xspace}
\newcommand{\cbtfc}{\textsc{CBT-FC}\xspace}
\newcommand{\cbtdp}{\textsc{CBT-DP}\xspace}

\title{\textsc{CBT-Bench}: Evaluating Large Language Models on Assisting \\ Cognitive Behavior Therapy}
\author{\textbf{Mian Zhang}\textsuperscript{*}\textsuperscript{1}, 
\textbf{Xianjun Yang}\textsuperscript{*}\textsuperscript{2}, 
\textbf{Xinlu Zhang}\textsuperscript{2}, 
\textbf{Travis Labrum}\textsuperscript{3}, \\
\textbf{Jamie C. Chiu}\textsuperscript{4},
\textbf{Shaun M. Eack}\textsuperscript{3}, 
\textbf{Fei Fang}\textsuperscript{5},
\textbf{William Yang Wang}\textsuperscript{2},
\textbf{Zhiyu Zoey Chen}\textsuperscript{1}
\\
  \textsuperscript{1}Department of Computer Science, The University of Texas at Dallas, \\
  \textsuperscript{2}Department of Computer Science, University of California, Santa Barbara, \\
  \textsuperscript{3}School of Social Work, University of Pittsburgh, \\
  \textsuperscript{4}Department of Psychology, Princeton University, \\
  \textsuperscript{5}School of Computer Science, Carnegie Mellon University \\
  {\tt \{mian.zhang, zhiyu.chen2\}@utdallas.edu}, \tt xianjunyang@cs.ucsb.edu}

\begin{document}
\maketitle

{\let\thefootnote\relax\footnotetext{\hspace{-1.5mm}*\hspace{0.2mm}Equal contributions.}}

\begin{abstract}
There is a significant gap between patient needs and available mental health support today. In this paper, we aim to thoroughly examine the potential of using Large Language Models (LLMs) to assist professional psychotherapy. To this end, we propose a new benchmark, \data, for the systematic evaluation of cognitive behavioral therapy (CBT) assistance. We include three levels of tasks in \data: \textbf{\RN{1}: Basic CBT knowledge acquisition}, with the task of multiple-choice questions; \textbf{\RN{2}: Cognitive model understanding}, with the tasks of cognitive distortion classification, primary core belief classification, and fine-grained core belief classification; \textbf{\RN{3}: Therapeutic response generation}, with the task of generating responses to patient speech in CBT therapy sessions.
These tasks encompass key aspects of CBT that could potentially be enhanced through AI assistance, while also outlining a hierarchy of capability requirements, ranging from basic knowledge recitation to engaging in real therapeutic conversations. We evaluated representative LLMs on our benchmark. Experimental results indicate that while LLMs perform well in reciting CBT knowledge, they fall short in complex real-world scenarios requiring deep analysis of patients' cognitive structures and generating effective responses, suggesting potential future work.\footnote{The data and code are released at \url{https://github.com/mianzhang/CBT-Bench}.}
\end{abstract}
\section{Introduction}
Mental health conditions have reached alarming levels globally, with one in eight people affected, according to the World Health Organization (2023)\footnote{https://www.who.int/campaigns/world-mental-health-day/2023}. There is a severe shortage of mental health professionals. In the U.S., more than 160 million people live in areas with insufficient mental health providers, with rural regions being especially underserved\footnote{https://bhw.hrsa.gov/data-research/review-health-workforce-research}. This critical gap underscores the need for AI-driven tools to support professionals and expand access to care.
Existing research has explored mental health condition classifications~\cite{gao2018machine,senn2022ensembles}, empathetic conversations~\cite{sharma2021towards,sharma2023human,adikari2022empathic}, and chatbots designed for simple discourse structures~\cite{hsu2023helping}.
However, work on professional assistance in real therapy settings remain limited. 
Some studies have addressed specific tasks in cognitive behavioral therapy (CBT)~\cite{beck2020cognitive}, such as cognitive distortion classification~\cite{chen2023empowering}. 
However, there are many other critical stages in CBT that could potentially be enhanced through automation.

In this work, we aim to thoroughly investigate the proficiency and potential of LLMs in supporting various facets and stages of professional mental health care. To this end, we propose \data, a systematic benchmark for evaluating CBT efficacy. \data is structured in three levels, from CBT knowledge recitation to the therapeutic responses generation in CBT sessions, providing a hierarchical assessment of CBT capabilities. To ensure the professionalism and high quality of our benchmark, we collaborate with domain experts (clinical psychologists, professors, and social workers) throughout the construction of \data.

In \textbf{level \RN{1}}, we aim to assess \textbf{basic CBT knowledge acquisition}. We propose \textsc{CBT-QA}, a new dataset of 220 multiple-choice questions. The QA pairs are collected from CBT exam questions for Master of Social Work (MSW) students as well as compositions from CBT experts, covering a wide range of CBT knowledge, including basic concepts, practical knowledge, and case studies. 

In \textbf{level \RN{2}}, we aim to assess \textbf{cognitive model understanding}. Modeling how the cognitive components of the patients, such as the beliefs and thoughts, are connected, is at the core of CBT~\cite{beck2020cognitive,kuyken2011collaborative}. 
In this work, we propose \textsc{CBT-CD}, a new dataset of cognitive distortion classification with 146 high-quality examples. 
We also propose two new tasks to assess the understanding of patients' beliefs: primary core belief classification and fine-grained core belief classification. For primary core belief classification, we construct \cbtmc, a new dataset of 184 examples with three primary core belief categories. For fine-grained core belief classification, we propose \cbtfc, a new dataset of 112 examples with nineteen fine-grained core belief categories. Building AI models for such tasks to assist the cognitive modeling process has the great potential to enhance therapists' accuracy and productivity.

In \textbf{level \RN{3}}, we aim to evaluate \textbf{therapeutic response generation}, the ultimate CBT efficacy -- whether the model can effectively respond to patient speech during CBT sessions like the therapists. Due to the privacy constraints associated with real CBT session data, accessing extensive datasets is challenging. In collaboration with professors specialized in CBT education, we propose \cbtdp, using \textit{Deliberate Practice} as outlined in \citet{boswell2022deliberate} for our assessment. 
This methodology is traditionally employed in assessing CBT proficiency among graduate students. It encompasses a collection of exercises to respond to typical patient speeches across ten key aspects of CBT sessions, categorized into three levels of difficulty, totaling 156 distinct exercises. This approach ensures comprehensive coverage of critical and challenging scenarios likely to occur in real sessions.
Generating high-quality responses to patient speeches in \cbtdp can serve as an effective proxy for real CBT session efficacy. 

We experiment with six popular LLMs for level \RN{1} and level \RN{2} in \data and find that 1) the models of large sizes are better at answering CBT knowledge questions; 2) simply making the models larger could not enhance their understanding ability of the cognitive model; 3) current LLMs struggle with detecting fine-grained cognitive disorders or core beliefs. A deeper analysis of the model performance for each class points out the main reasons for a wrong prediction and directions to improve the corresponding ability of the models. 
For the level \RN{3} task, we find that LLMs generally follow a rigid logical reasoning process but lack a crucial skill in psychotherapy—thinking and guiding from the patient’s perspective to respect their autonomy and build rapport. These findings highlight significant limitations in applying LLMs to real-world psychotherapy practice and provide valuable insights for future research and development toward more accessible and efficient mental health care.

\section{Related Work}
Our work is the first to systematically evaluate LLMs' ability to assist professional human therapists in the specialized field of CBT. The works directly related to our research include injecting domain knowledge of mental health into the models~\cite{Yang2024-rp,Kim2024-xp}, cognitive disorder detection~\cite{Shreevastava2021-eh,Wang2023-si,chen2023empowering}, negative thoughts recognition and reframing~\cite{Sharma2023-dj,Maddela2023-yb,Sharma2024-jy}, and patient simulation~\cite{Chen2023-dk} or therapist simulation~\cite{Liu2023-ub} in therapeutic conversation. \citet{wang2024patient} proved that with clear modeling of the cognitive model of patients, LLMs could act more like real patients. Also, we believe that enhancing the understanding of the cognitive models of patients could be beneficial to interpretable medical decisions~\cite{Yang2023-mp}, which is a crucial step towards reliable and safe automated mental healthcare~\cite{Ji2023-gn,Grabb2024-vq}. In the proposed therapeutic response generation task, we leverage the feedback from professional therapists to assess the potential of LLMs responding like a therapist in real therapy sessions, while~\citet{Louie2024-kj} explored how to make LLMs roleplay patients with the feedback from professional therapists.

\section{\data}
In this section, we elaborate on how \data is constructed. We discuss  \textbf{level \RN{1}, \RN{2}, and \RN{3}} tasks in \S\ref{data_1}, \S\ref{data_2}, and \S\ref{data_3}, respectively. Note that apart from collaborating with social work professors and clinical psychologists, who are our co-authors, we recruit CBT experts (clinical psychologists, social workers, etc.) from UpWork\footnote{www.upwork.com} for all additional data annotation tasks (See \S\ref{ethics} for recruiting details).

\subsection{Level \RN{1}: Basic CBT Knowledge Acquisition}
\label{data_1}
\begin{table*}[t]
\small
\begin{center}
\resizebox{\textwidth}{!}{
\begin{tabular}{lll}
\toprule
\textbf{Knowledge Types} & \textbf{Example QA Pairs from \cbtqa} & \textbf{Distributions (\%)} \\
\midrule
Basic CBT knowledge and concepts & \makecell[l]{Albert Ellis’ Cognitive Model includes which components? \\ \textbf{A. }Activating Events – Behaviors – Cognitions, \textbf{B. }Antecedents – Beliefs – Consequences, \\ \textbf{C: }Activating Events – Beliefs – Consequences, \textbf{C. }Antecedents – Behaviors – Consequences} & 41.82 \\
\midrule
Practical CBT knowledge & \makecell[l]{When helping clients evaluate automatic thoughts, \\ therapists should generally help clients evaluate which aspects of those thoughts? \\ \textbf{A. }Accuracy and/or intensity, \textbf{B. }Intensity and/or utility, \\ \textbf{C. }Accuracy and/or utility} & 34.09 \\
\midrule
Case studies & \makecell[l]{The client has identified an automatic thought of "My partner is going to break up with me". \\ The therapist asks the client, "If your thought is accurate and your partner does break up with you, \\ what does that mean about you?". The therapist is most likely trying to identify: \\\textbf{A. }The client's intermediate belief, \textbf{B. }The client's core belief, \\ \textbf{C. }The client's thinking error } & 18.18\\
\midrule
Others & \makecell[l]{What are some ways that CBT therapists can engage in therapy from a multicultural perspective? \\ (select all that apply) \\ \textbf{A. }Not take clients from a different culture than their own,  \\ \textbf{B. }Ask clients about the strengths and challenges of their cultural, racial, and ethnic identity during intake, \\ \textbf{C. }Being aware of their own cultural values and biases, \\ \textbf{D. }Work together with the client to incorporate the client's core values} & 5.9\\
\bottomrule
\end{tabular}
}
\caption{Knowledge types in \cbtqa, with example QA pairs and distributions in the test set. }
\label{table:cbt_qa}
\end{center}
\end{table*}
To evaluate the basic CBT knowledge acquired by LLMs, we propose a new dataset, \cbtqa, encompassing 220 multiple-choice questions. We first worked with two social work professors and collected 92 multiple-choice questions from the exams they issued for graduate CBT courses. Each question has 2-5 answer choices. Then, we hired four CBT experts to compose new QA pairs, using the QA pairs from the exam source as guidance, and each expert was tasked with composing 50. To ensure high quality, we required the experts to cross-verify the QA pairs and excluded those deemed of low quality. We ended up with 178 high-quality QA pairs, which, when combined with the 92 ones sourced from exams, resulted in our \cbtqa dataset of 270 QA pairs. We randomly selected 50 ones for use such as in-context learning examples. The remaining 220 pairs were designated as the final test set.

To get a better understanding of the fine-grained knowledge types involved in \cbtqa, we employed another two CBT experts to categorize the test set into four categories. Table~\ref{table:cbt_qa} shows example QA pairs from the four categories and corresponding distributions in the test set. See Appendix~\ref{sec:apx-example} for more examples. To assess human performance, we employed another two CBT experts to solve the test set and ended up with an average accuracy of 90.7\% with an agreement rate over 80\%.

\subsection{Level \RN{2}: Cognitive Model Understanding}
\label{data_2}
\begin{table*}[t]
\scriptsize
\begin{center}
\resizebox{0.95\textwidth}{!}{
\begin{tabular}{lp{12cm}p{3cm}}
\toprule
\textbf{Datasets} & \textbf{Inputs} & \textbf{Labels} \\
\midrule
\cbtcd & \textbf{Situation}: Our wedding was put off because his parents asked him to build a house for them 2 months before our wedding! They had a perfectly good house at the time they just wanted their dream house. \textbf{Thoughts}: I am a victim with no power in this situation. I must accept this behavior. I am too scared to leave this situation. I am not worthy of better. His parents hate me. His parents do not want us to get married. He may not want to marry me either. He loves his parents more than me. I will always be second in his life. His parents had no need for a house, and I know this for sure. I am aware of every aspect of this situation. & all-or-nothing thinking; personalization; mind reading \\
\midrule
\cbtmc & \textbf{Situation}: I had an amazing childhood. When I was twelve in 2004, my father had to go to Iraq. My mother thought it would be best if she moved my brother and I back to the U.S., where we would have family support. I was very depressed because my dad was my hero and I blamed my mom for everything that went wrong. I felt like no one understood me... When my dad came back, he wanted a divorce from my mother. \textbf{Thoughts}: Everything was great until my mom messed everything up. Because of her, my brother and I had to leave our dad. We ended up living somewhere where no one liked me. It's her fault that I never felt like I fit in. Even when my dad came back, he didn't want us either - he wanted a divorce. I guess no one will ever want me in their life. I'll probably be alone forever. & helpless; unlovable\\
\midrule
\cbtfc & \textbf{Situation}: My daughter was recently diagnosed as bipolar. If I say anything about seeking treatment, my daughter accuses me of not understanding her and what is happening to her. She is very paranoid and worries about her safety all the time. I need to know how to talk to her and what to do to get her into treatment. \textbf{Thoughts}: I am a bad mother. This is my fault. It is so shameful that my daughter has bipolar. If my daughter gets worse, then it will be my fault for not getting her into treatment. I need to do something. This is my responsibility. & I am incompetent; I am helpless; I am powerless, weak, vulnerable; I am bad - dangerous, toxic, evil \\
\bottomrule
\end{tabular}
}
\caption{Examples from \cbtcd (cognitive distortion classification), \cbtmc (primary core belief classification), and \cbtfc (fine-grained core belief classification). See more examples in Appendix~\ref{sec:apx-example}. }
\label{table:cbt_level2_examples}
\end{center}
\end{table*}

\begin{table}[t]
\small
\begin{center}
\resizebox{0.46\textwidth}{!}{
\begin{tabular}{lccc}
\toprule
 & \cbtcd & \cbtmc & \cbtfc \\
\midrule
\# of examples & 146 & 184 & 112 \\
\midrule
\# of labels & 10 & 3 & 19 \\
\midrule
Average situation length & 232.9 & 240.7 & 233.4 \\
\midrule
Average thought length & 258.8 & 256.9 & 248.4 \\
\midrule
Average ground truth labels & 2.5 & 1.9 & 3.8 \\

\bottomrule
\end{tabular}
}
\caption{Statistics of three level \RN{2} tasks.}
\label{table:cbt_level2}
\end{center}
\end{table}
In the exam setting of \cbtqa, most questions assess the models' ability to recite knowledge, as LLMs may have seen this information during pre-training—especially when it comes to basic CBT knowledge and concepts that are prevalent online. However, knowledge recitation alone is insufficient to support the effective use of LLMs in assisting real practice. To address this, we propose the Level \RN{2} task set: cognitive model understanding. In CBT, therapists develop a cognitive model to represent a patient's unhealthy cognitive processes that contribute to their mental health issues~\cite{beck2020cognitive}. A cognitive model typically includes beliefs, thoughts, emotions, and other relevant elements. The associations among these components provide a clear representation of the patient's maladaptive cognitive process (See Appendix~\ref{sec:apx-example} for an example cognitive model).
Constructing a patient’s cognitive model is a crucial step in CBT~\cite{beck2020cognitive,beck2014advances,wang2024patient}, as therapists must first summarize these maladaptive structures before working to correct them. We believe that the process of cognitive model construction can be effectively enhanced with the assistance of LLMs, which could help in automatically identifying key components—such as beliefs and thought patterns—and their associations.

Existing work mostly focuses on cognitive distortion classification~\cite{sharma2023cognitive,shreevastava2021detecting,chen2023empowering}--classifying the maladaptive thinking patterns in a patient's speech, which can be viewed as one component of cognitive modeling. However, existing datasets are typically limited by poor quality~\cite{shreevastava2021detecting,chen2023empowering} or simplicity of the input speech. In our Level \RN{2} tasks, we first propose a new dataset, \cbtcd, for cognitive distortion classification with enhanced quality. We use the label set in~\cite{shreevastava2021detecting} of ten distortion types.

We then propose two new tasks centered on core beliefs. Core beliefs are at the center of a patient’s cognitive model—deeply ingrained perceptions about oneself, others, and the world~\cite{beck2020cognitive}. Eliciting core beliefs is one of the most challenging stages of cognitive modeling, while the essential focus in later sessions is to correct such unhealthy, biased core beliefs. According to \citet{beck2020cognitive}, there are three primary core beliefs: \textit{helpless, unlovable,} and \textit{worthless}. Each of these categories encompasses more specific, fine-grained core beliefs. For example, \textit{I am incompetent} and \textit{I am a loser} fall under \textit{helpless}. In total, these fine-grained core beliefs are classified into 19 categories. In our Level \RN{2} tasks, we propose new datasets \cbtmc for primary core belief classification, and \cbtfc for fine-grained core belief classification. Check Appendix~\ref{sec:apx-label} for detailed label definitions and distribution for all three datasets. 

We use the patient speech in the TherapistQA dataset~\cite{shreevastava2021detecting}, sourced from mental health-related posts in online forums, as our data source to construct \cbtcd, \cbtmc, and \cbtfc. Our annotation process is as follows: (1) From each original post, we instruct the expert annotators to extract a segment that describes a situation along with the associated thoughts. Annotators may also create plausible imagined situations and thoughts to supplement the original segments if they are insufficiently detailed. (2) For cognitive distortions, we instruct the experts to annotate up to three key ones. For primary core beliefs and fine-grained core beliefs, we instruct the experts to annotate up to three primary beliefs. Under each identified primary core belief, we annotate up to three fine-grained beliefs. 

We employed a total of six CBT experts from UpWork to conduct the annotation. Check Appendix~\ref{sec:ann_details} for our annotation interface. We conducted another round of cross-verification to filter out the low-quality ones. We end up with 146 examples for \cbtcd, 184 examples for \cbtmc, and 112 examples for \cbtfc. We also have another 20 examples for each task serving as other potential uses. Table~\ref{table:cbt_level2_examples} and \ref{table:cbt_level2} presents some statistics and examples of the three datasets. We employ two new experts to perform another round of annotation for each dataset to establish human expert performance references. We ended up with a weighted F-1 score of 49.1\%, 77.7\%, and 54.6\% for \cbtcd, \cbtmc, and \cbtfc, respectively. Despite implementing a cross-verification step to filter out low-quality data, the results suggest that these tasks inherently involve a degree of subjectivity.

\subsection{Level \RN{3}: Therapeutic Response Generation}
\label{data_3}
\begin{table*}[t]
\small
\begin{center}
\resizebox{\textwidth}{!}{
\begin{tabular}{lccc}
\toprule
\textbf{Category} & \textbf{Difficulty Level} & \textbf{Patient Speech} & \textbf{Reference Response} \\
\midrule
Negotiating a session agenda & Beginner & \makecell[l]{[Nervous] I don’t think I’m ready for working on this today.} & \makecell[l]{OK. We can revisit and possibly modify our plan for today. How about \\ we first take a step back and explore your thinking about this? What \\ thoughts are you noticing as we discuss
the agenda?} \\
\midrule
Negotiating a session agenda & Intermediate & \makecell[l]{[Agitated] Wow. You won’t believe what happened this week. \\ It’s a really long story . . .} & \makecell[l]{It sounds like there’s a lot on your mind, and I’d like to hear about it. \\ Would it be OK to take a second to discuss our agenda for the day first, \\ including where discussing this past week might fit in, as well as anything \\ else you want to take up here today? I want to make sure that we budget \\ our time accordingly. Shall we start with a homework check-in and then \\ 
tackle the story?} \\
\midrule
\makecell[l]{Responding to therapeutic \\ alliance ruptures} & Advanced & \makecell[l]{[Anxious] I did the measure you asked me to fill out. Honestly, \\ I think I might have rated you lower than usual, but I’m not sure \\ I want to talk about it.} & \makecell[l]{I was just noticing that your trust in me has gone down some. I wonder \\ if you could help me appreciate what that’s like for you? I’d far prefer \\ that to just persisting with our agenda when you may have diminishing \\ faith in it or me.} \\
\bottomrule
\end{tabular}
}
\caption{Three example deliberate practice exercises from \cbtdp. The patient speech in each exercise starts with an emotion label indicating the patient's emotional status. }
\label{table:cbt_level3}
\end{center}
\end{table*}

At this level, we take a further step to explore a longstanding and challenging question: \textit{\textbf{Do LLMs have the ability to conduct effective therapeutic conversations?}} This ultimate capability assesses CBT competence as a whole, beyond the specific tasks in levels \RN{1} and \RN{2}. In real practice, we must proceed with caution, using such applications to assist therapists rather than interacting directly with patients without supervision. Nonetheless, first and foremost, it is crucial to understand the LLMs' capability in providing effective responses.

Collecting a sufficient amount of real CBT session data is difficult due to privacy constraints. Taking the suggestions from professors specialized in CBT education (our co-authors), we propose using \textit{Deliberate Practice}~\cite{boswell2022deliberate} as our level \RN{3} task, \cbtdp. This methodology is traditionally used to assess graduate students' proficiency in CBT. Unlike simulated scenarios such as role-playing, deliberate practice involves a structured set of exercises focusing on responding to typical patient dialogue in CBT sessions. These exercises are divided into ten categories, such as  \textit{Working With Cognitions} and \textit{Responding to Client Resistance}, addressing key aspects and challenges of conducting CBT. This approach ensures comprehensive coverage of critical and difficult situations likely to arise in real therapy. Generating high-quality responses to patient speech in \cbtdp can serve as an effective proxy for evaluating real-world CBT session efficacy.

Each category of \cbtdp contains approximately 15 exercises across three difficulty levels: beginner, intermediate, and advanced. We use the 156 exercises outlined in \citet{boswell2022deliberate} as \cbtdp, where the authors provide one reference response for each exercise. Table~\ref{table:cbt_level3} shows some example exercises with reference responses. Check Appendix~\ref{sec:apx-label} for all categories and statistics of all exercises.
For model evaluations, we frame \cbtdp as a generation task. The patient's speech in each exercise serves as the input, and the model is expected to generate an appropriate response. We work with our co-authors (professors in social work) to propose a set of evaluation criteria, four fine-grained criteria for each category (Appendix~\ref{sec:apx-label}). As suggested by our co-authors, while \citet{boswell2022deliberate} provide a set of reference responses, these should not be treated as the sole `gold standard.' Consequently, matching-based automatic evaluation metrics may not provide an accurate measure for this task. Therefore, we employ experts to conduct pairwise comparisons between the model generation and reference response under each criterion.

\section{Level \RN{1} and \RN{2} Experiments}
\subsection{Settings}
We evaluated six popular LLMs' performance on basic CBT knowledge acquisition (level \RN{1}) with \cbtqa and cognitive model understanding (level \RN{2}) with \cbtcd, \cbtmc, and \cbtfc. The models include a close-sourced model GPT-4o\footnote{https://openai.com/index/gpt-4o-system-card/} and five open-sourced models of sizes ranging from 7B to 405B: Mistral-v0.3-7B~\cite{Jiang2023-hu}, Gemma-2-9B~\cite{Gemma-Team2024-jc}, Llama-3.1-8B, Llama-3.1-70B, and Llama-3.1-405B~\cite{Llama-Team2024-vb}\footnote{We always use the instruct version of the open-sourced models for all the experiments in this paper.}. We refer to models with sizes under 10B as small models while others as large models (including GPT-4o). 

We cast the classification tasks of level \RN{2} into multiple-choice questions, where each question can have more than one option. Prompt examples for each task of level \RN{1} and \RN{2} can be found in Appendix~\ref{sec:task-prompts}. The inferencing temperature is set as 0.0 to eliminate the randomness of the model generations for reproducibility and data type bfloat16 is used for the model weights and activations. 

\subsection{Results}
\begingroup
\small
\setlength{\tabcolsep}{4pt}
\begin{table*}[h]
    \centering
    \small
    \begin{tabular}{l r rrr rrr rrr}
        \toprule
           & \multicolumn{1}{c}{\cbtqa} & \multicolumn{3}{c}{\cbtcd} & \multicolumn{3}{c}{\cbtmc} &       \multicolumn{3}{c}{\cbtfc} \\
            \cmidrule(lr){2-2} \cmidrule(lr){3-5} \cmidrule(lr){6-8} \cmidrule(lr){9-11}
            \textbf{Model} & Accuracy & Precision & Recall & F1 & Accuracy & Recall & F1 & Accuracy & Recall & F1 \\
            \midrule
            Mistral-v0.3-7B & 75.9 & 43.8 & 56.4 & \textbf{45.3} & 78.8 & 75.0 & 72.5 & 43.8 & 59.7 & 47.9 \\
            Gemma-2-9B & 90.0 & 44.5 & 47.2 & 40.1 & 79.5 & 82.9 & 80.4 & 45.4 & 63.3 & 51.5 \\
            Llama-3.1-8B & 78.2 & 39.7 & 59.9 & 44.1 & 75.2 & 93.9 & \textbf{82.5} & 37.0 & 62.8 & 44.8 \\
            Llama-3.1-70B & 92.7 & 50.9 & 51.8 & 43.9 & 82.6 & 72.7 & 75.8 & 62.0 & 56.6 & 55.4 \\   
            Llama-3.1-405B & \textbf{95.0} & 49.4 & 44.2 & 43.3 & 85.3 & 70.1 & 75.3 & 53.1 & 68.7 & \textbf{58.2} \\  
            GPT-4o & 94.1 & 55.8 & 52.0 & 43.9 & 80.2 & 77.6 & 78.4 & 54.5 & 62.1 & 56.7 \\            
          \midrule
          Human & 90.7 & 51.1 & 48.1 & 49.1 & 76.0 & 79.4 & 77.6 & 53.0 & 57.3 & 54.6 \\
         \bottomrule
    \end{tabular}
    \caption{Performance of LLMs on basic CBT knowledge acquisition (\cbtqa) and cognitive model understanding (\cbtcd, \cbtmc, and \cbtfc). The precision, recall, and F1 are averaged by class portion.}\label{tab:level12_main}
\end{table*}
\endgroup

Table~\ref{tab:level12_main} shows the performance of the LLMs. We report accuracy for \cbtqa and weighted precision, recall, and F1 score for the other three datasets. Here are the findings:
\begin{figure*}[htbp]  
    \centering  
    \begin{subfigure}[b]{0.45\textwidth}
        \centering
        \includegraphics[width=\textwidth]{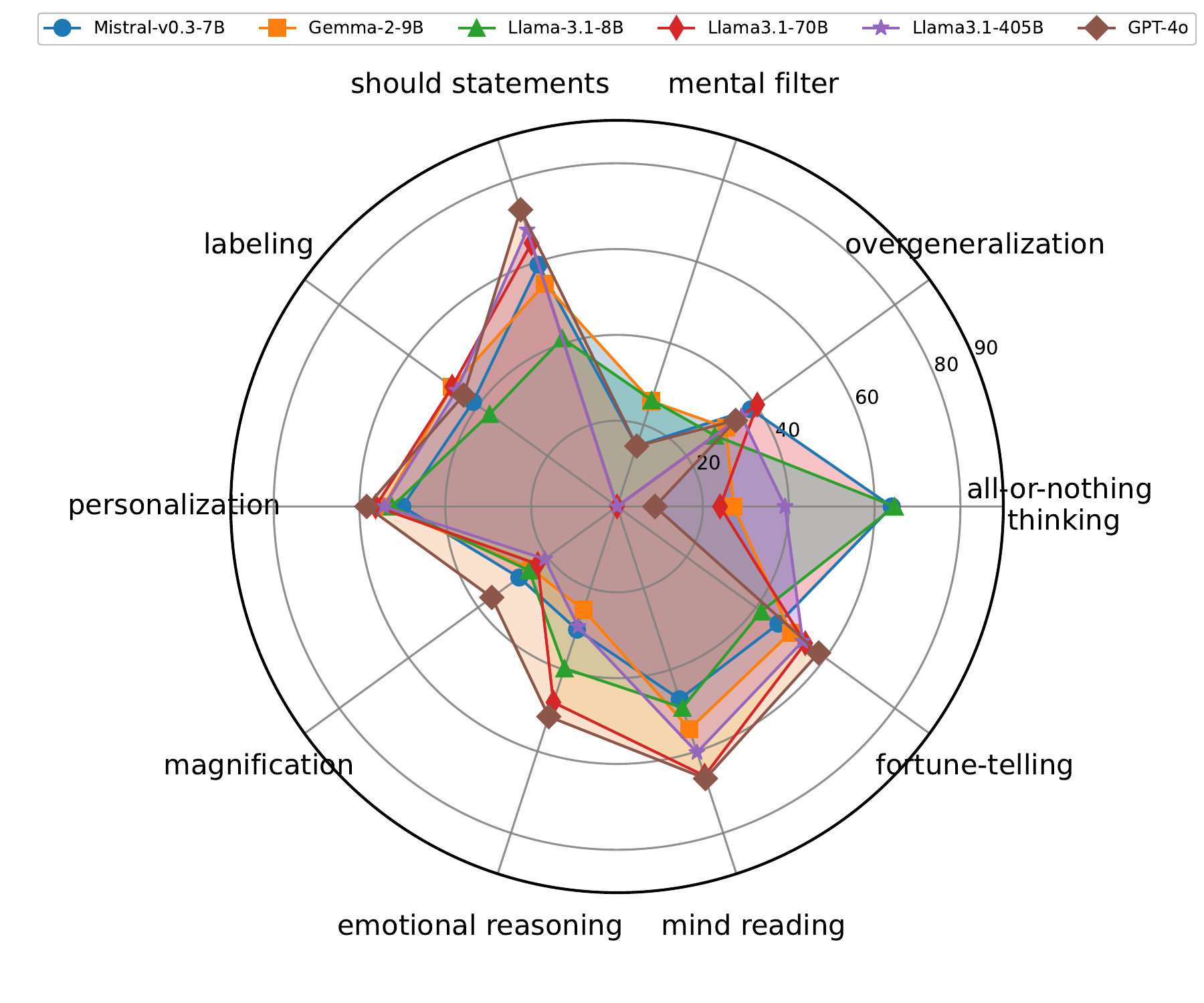}
        \caption{\cbtcd}
        \label{subfig:task2}
    \end{subfigure}
    \hspace{0.02\textwidth}
    \begin{subfigure}[b]{0.45\textwidth}
        \centering
        \includegraphics[width=\textwidth]{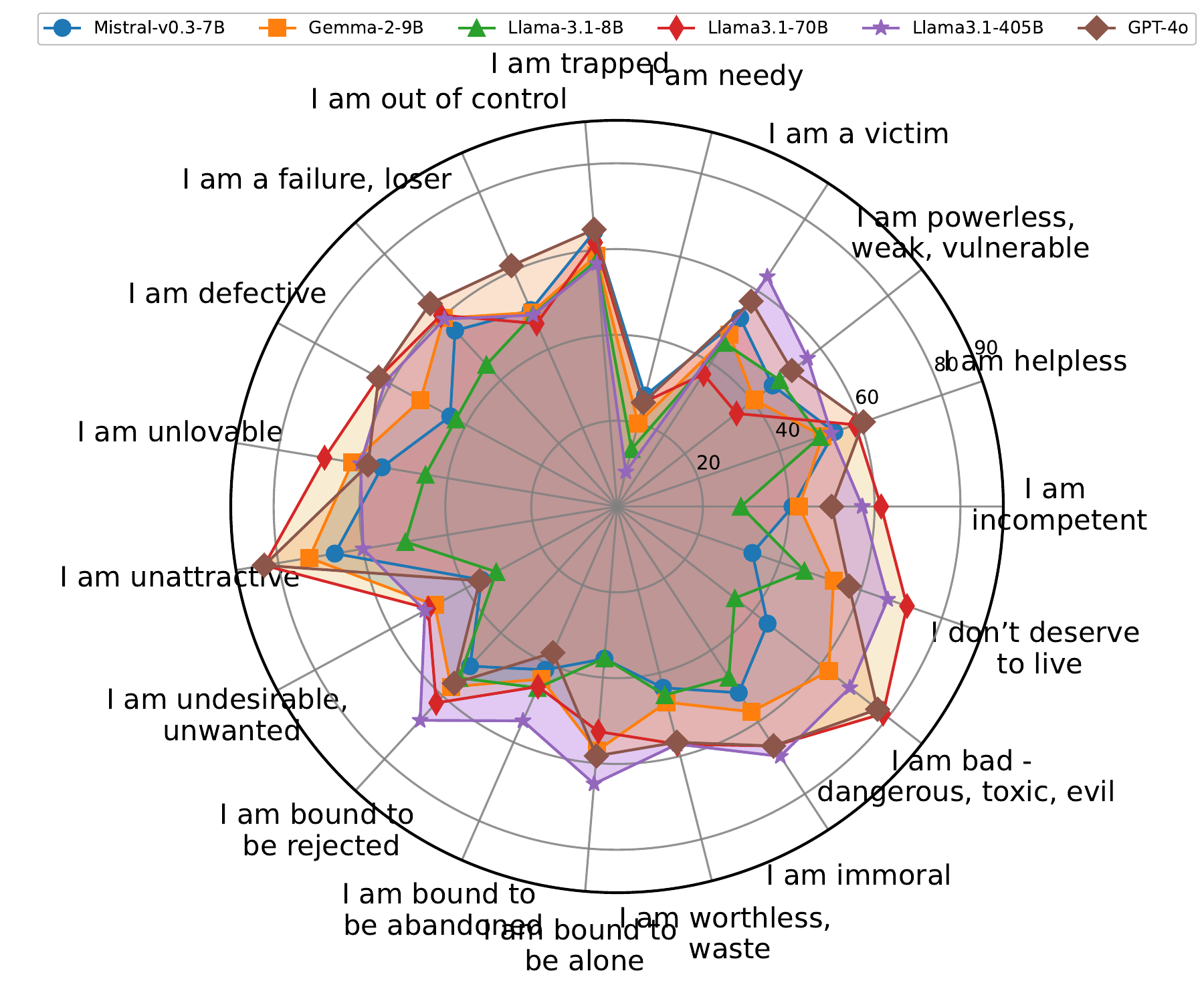}
        \caption{\cbtfc}
        \label{subfig:task4}
    \end{subfigure}

    \caption{Detailed F1 scores of each label for \cbtcd and \cbtfc.}
    \label{fig:class-metric}
\end{figure*}

\noindent \textbf{Large LLMs are better at answering CBT knowledge questions.} Large models could achieve higher accuracies on \cbtqa than small models. This could be attributed to larger models storing more knowledge. However, Gemma-2-9B has a similar model size as Llama-3.1-8B and Mistral-v0.3-7B and it surpasses these two by a significant margin, even comparable to Llama3.1-70B. This may be caused by the distribution difference of the data used for training.

\noindent \textbf{Simply making the models larger could not enhance their understanding ability of the cognitive model.} Mistral-v0.3-7B and Llama-3.1-7B these two small models get the best F1 scores on \cbtcd and \cbtmc, outperforming models of large sizes. This gives us insight that when increasing the model size to pursue better general capabilities like reasoning, the expertise of mental health care should also receive attention.

\noindent \textbf{Current LLMs struggle with detecting fine-grained cognitive disorders or core beliefs.} \cbtcd and \cbtfc are very challenging for current LLMs. The models generally perform poorly on these two datasets which are very difficult even for professional therapists. Enhancing these abilities of LLMs could make the therapy process more efficient and accurate, and our datasets serve as valuable resources to propel potential advances.

To delve further, we examine the accuracies for questions of different knowledge types in \cbtqa and detailed F1 scores on labels in \cbtcd, \cbtmc, and \cbtfc. The results for two fine-grained classification tasks, \cbtcd and \cbtfc, are shown in Figure~\ref{fig:class-metric}\footnote{The results for \cbtqa and \cbtmc are shown in Appendix~\ref{sec:apx-class-metrics} for space limit.}. We could see that for \cbtqa, questions of \textit{basic CBT knowledge and concepts} and \textit{others} are easier to answer than questions of \textit{case studies} and \textit{practical CBT knowledge}. This could be the reason that questions of \textit{case studies} and \textit{practical CBT knowledge} require more understanding of how to use the corresponding knowledge, not just reciting them. For \cbtcd, the models all struggle to identify \textit{mental filter}, \textit{manification}, and \textit{overgeneralization} disorders. And large models are not consistently outperforming small models. For \textit{all-or-nothing thinking} disorder, two small models have the best F1 scores. And for \textit{emotional reasoning} disorder, GPT-4o and Llama3.1-70B have the best performance while Llama3.1-405B even falls behind Llama3.1-7B. For \cbtmc, small models are better for \textit{helpless} core belief, and large models take the lead for \textit{worthless}. For \textit{unlovable} belief, their performance is not much different. For \cbtfc, models are struggling with detecting \textit{I am needy} core belief.

We go through the prediction of Llama-3.1-405B on the two fine-grained classification datasets and present some typical ones in Appendix~\ref{sec:error-ans-level-12}. The main reasons for a wrong prediction include: 1) the model is not sensitive to the indicators of cognitive disorders or core beliefs, such as "People do think I'm..." of \textit{overgeneralization} disorder or "There is nothing enjoyable in my life" of \textit{I don’t deserve to live} core belief; 2) The model focuses on several disorders or beliefs while neglecting the others; and 3) The model makes a decision without sufficient supporting conditions. For example, determining the \textit{personalization} disorder can be divided into one attributing the negative behavior to some person and this person should be himself; one of the cases in \cbtfc only meets the first but is judged as \textit{personalization} by the model.

\section{Level \RN{3} Experiments}

\begin{table*}[h!]
\centering
\resizebox{0.8\textwidth}{!}{
\begin{tabular}{cccccccccccc}
\toprule
\small{Exercise} & \small{1} & \small{2} & \small{3} & \small{4} & \small{5} & \small{6} & \small{7} & \small{8} & \small{9} & \small{10} & \small{Avg.} \\ \hline
\small{Llama-3.1-405B }  & \small{0.07} & \small{0.06} & \small{0.21} & \small{-0.24} & \small{-0.19} & \small{0.00} & \small{0.18} & \small{-0.31} & \small{0.07} & \small{0.00} & \small{-0.01} \\  
\small{Llama-3.1-8B }   & \small{-0.21} & \small{-0.31} & \small{-0.47} & \small{-0.29} & \small{-0.13} & \small{0.00} & \small{-0.35} &  \small{0.00} & \small{-0.13} & \small{-0.33} & \small{-0.22} \\  
\small{GPT-4o }  & \small{-0.50} & \small{-0.50} & \small{-0.13} & \small{-0.24} & \small{-0.31} & \small{-0.53} & \small{-0.06} &  \small{-0.44} & \small{-0.13} & \small{-0.40} & \small{-0.32} \\
\bottomrule
\end{tabular}
}
\caption{Overall rating: pairwise comparison of different models vs. reference across ten exercises. }\label{sep3: overall}
\end{table*}

\begin{table*}[ht]
\centering
\small
\resizebox{0.98\textwidth}{!}{
\begin{tabular}{p{3cm}p{15cm}}
\toprule
\textbf{Statement} & [Frustrated] I know I said I would go to the event and start a conversation with new people, but I only 
managed to say “hello” to one person. \\
\midrule
\textbf{Llama-3.1-405B} & It sounds like you're feeling disappointed with yourself for not meeting your own expectations. Can we take a closer look at what held you back from starting more conversations? Was it anxiety, fear of rejection, or something else? Let's break it down and identify some smaller, achievable steps for next time. \\
\midrule
\textbf{Human Reference} & It seems like you might be a bit disappointed by this, but I see this as significant progress. It was a big accomplishment to show up to the event in the first place and then say hello to someone you don’t know. Let’s talk about what that was like. \\
\bottomrule
\end{tabular}
}
\caption{Beginner exercise example: comparison between Llama-3.1-405B and human reference responses. The former is more structured, while the latter provides more emotional support.}\label{case_study_overall}
\end{table*}

\subsection{Settings}
We evaluate three LLMs' performance for \cbtdp (level \RN{3}): Llama-3.1-8B, Llama-3.1-405B, and GPT-4o.
For all generations, we keep the generation temperature at 0.7. For evaluation, as suggested by domain experts (\S\ref{data_3}), we conduct pairwise comparisons between model generations and the reference responses in \cite{boswell2022deliberate} under our proposed criteria set (\S\ref{data_3}). To ensure a fair comparison, we prompt the LLMs to generate responses within similar lengths as the reference (see Appendix~\ref{sec:task-prompts}).

\subsection{Expert Evaluation}
To minimize bias in the annotation process, we mixed and randomly shuffled the comparison pairs across all models. For each annotation pair, we randomized the order in which the model-generated and reference responses appeared in the annotation interface. This approach ensures that annotators remain unaware of the source of each response. 
We instruct the experts to label each pair with one of the following five labels: 1) A is much better than B, 2) A is slightly better than B, 3) A and B are about the same, 4) B is slightly better than A, and 5) B is much better than A. Labeling is done for each criterion defined, as well as an overall preference. We map the results to a scale from -2 to 2, where $+2$ indicates a strong preference for model-generated response.

The full annotation consists of 468 pairwise comparisons across all models, with each instance requiring approximately 3 minutes for the expert to finish. Given the expert's rate of \$60 per hour, the annotation process takes around 24 hours to complete, resulting in a significant cost. We first have our expert co-author annotate a sample and write explanations of scoring. Then, we hired another expert to finish the annotation of all data following the guidance of our co-author (Appendix~\ref{sec:ann_details} for our annotation interface).

\subsection{Overall Result and Analysis}
Here we first report the overall pairwise comparison results between each LLM vs. reference on 10 exercise categories in Table \ref{sep3: overall}. Notably, the average results show that all three LLMs still lag behind humans by a significant margin, with Llama-3.1-405B achieving the best result while GPT-4o is the worst. Moreover, Llama-3.1-8B and GPT-4o never surpass human-written references in every single class of exercise, while Llama-3.1-405B shows some superior results in half of the exercise categories. Further insights reveal that Llama-3.1-405B is better because it \textit{keeps engaging the client and asking pertinent questions to guide them}, but GPT-4o tends to \textit{correct the client's statement and keeps using professional CBT terminology to challenge the client}. And Llama-3.1-8B lies in the middle.
Although this might be improved by adding additional limits in the prompt, we believe the current prompt starting with "you are a professional therapist using cognitive-behavioral therapy (CBT) in a session with a client." should already inform the model about its role. The complete prompt can be found in Table \ref{tab:apx-prompt-task5}. We also draw the win-tie-loss figure in Appendix \ref{sec:win-tie}.

\begin{figure}[h]
\centering
\includegraphics[width=0.9\linewidth]{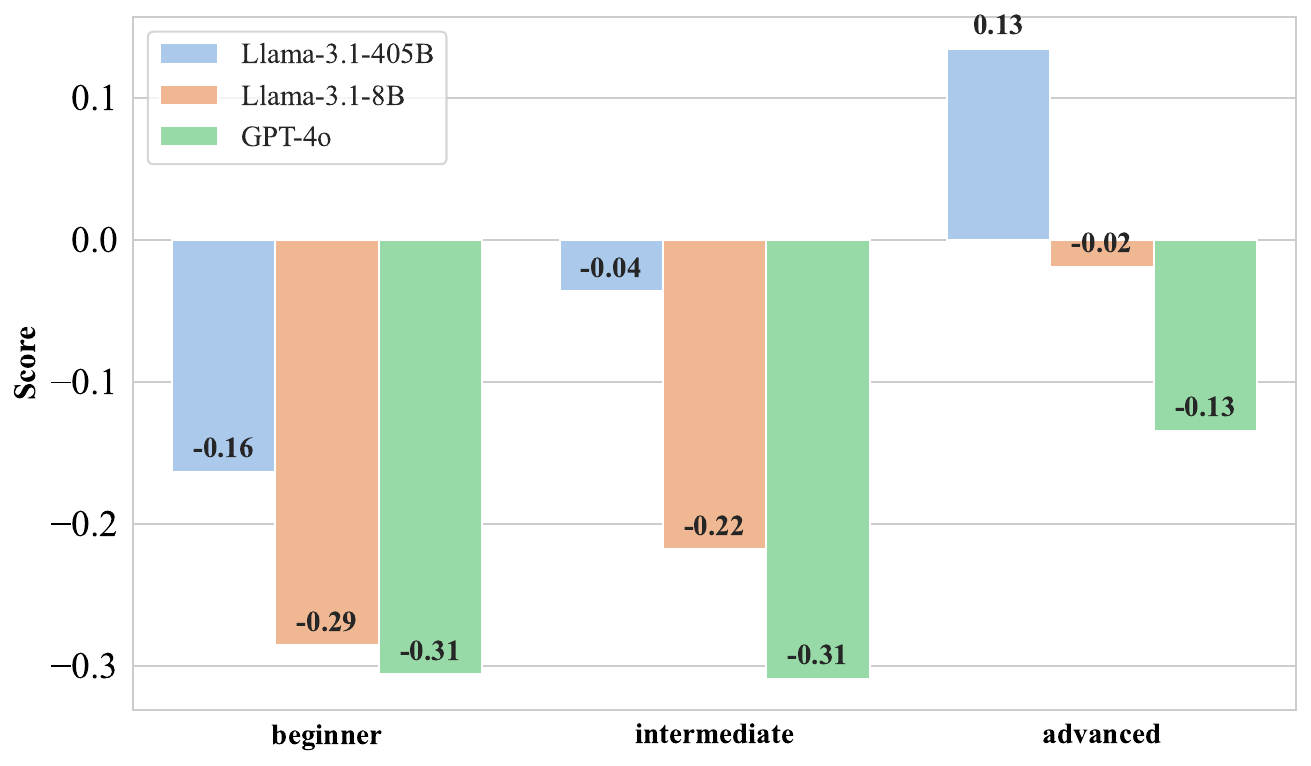}
\caption{The overall pairwise comparison of different models vs. reference across difficulty level.}\label{fig: difficulty_level}
\end{figure}

Next, we discuss the overall score categorized by difficulty level in Figure \ref{fig: difficulty_level}. Surprisingly, we find that the models generally perform better when the difficulty level increases. Using Llama-3.1-405B as an example, we observe that the model’s responses tend to be structured and rigid, often following a pattern such as repeating the client’s statement, questioning reasons, and proposing possibilities. In contrast, human reference responses are more effective in affirming the patient’s feelings and thoughts, and providing guidance \textit{from the patient’s perspective with a high degree of empathy and flexibility}, thereby establishing trust and rapport. At the beginner level, tasks tend to be logically simpler and more empathy-focused; consequently, model-generated answers are rated lower, as shown in Table \ref{case_study_overall}.

\begin{table}[h!]
\centering
\resizebox{0.45\textwidth}{!}{
\begin{tabular}{crrrr}
\toprule
\multirow{2}{*}{\small{\#Exe.}} & \multirow{2}{*}{\small{Metric}} & \multicolumn{3}{c}{\small{Model Results}} \\ \cline{3-5} 
 &  & \small{\makecell{Llama-3.1-405B}} & \small{\makecell{Llama-3.1-8B}} & \small{\makecell{GPT-4o}} \\ \hline
\multirow{4}{*}{\small{2}} 
 & \small{C1}  & \small{0.34} & \small{-0.25} & \small{0.25} \\ 
 & \small{C2}  & \small{0.06} & \small{0.13} & \small{-0.25} \\  
 & \small{C3}  & \small{-0.38} & \small{-0.13} & \small{-0.63} \\  
 & \small{C4}  & \small{0.56} & \small{0.13} & \small{0.81} \\  
 \bottomrule
\end{tabular}
}
\caption{Results breakdown of 4 criteria on questions from exercise 2, reported with the average score.}\label{sep3: c}
\end{table}

In addition, Table \ref{sep3: c} shows the models' scores across 4 different criteria (C1-4) for exercises 2 (full results for all exercises in Appendix~\ref{sec:apx-criteria-breakdown}). 
Apart from the overall rating, these criteria give fine-grained insights into the model's performance. 
We saw mixed results from the evaluation of exercise 2 (Establishing Goals). For example, in the comparison in appendix Table \ref{case_study: creteria}, criteria 1 is ``{Suggest CBT-consistent goals and tasks that align with an individualized CBT case formulation}'' and criteria 4 is ``{Emphasize concrete, actionable, and measurable goals}''. The human reference is much better in Criteria 1 and slightly worse in Criteria 4. Indeed, the model provides \textit{concrete and actionable suggestions (aligns with criteria 4) while the reference focuses on CBT case formulation (aligns with criteria 1)}.
However, for exercise 10 (Responding to Client Resistance), all criteria witness negative results on all models (see Appendix~\ref{sec:apx-criteria-breakdown}), consistent with the overall rating. The reason can be summarized as the models tend to \textit{dismiss the clients’ experience and challenge their resistance rather than being responsive to patient needs as human therapists}. We show two such examples in appendix Table \ref{case_study: e2_1} and \ref{case_study: e2_2}, clearly showing that the model tends to challenge the client while humans always respect clients' autonomy.

We provide a typical example comparing model responses in Table~\ref{case-study-level-3}. In this case, the client perceives the therapist’s suggestion of ‘little rewards’ as unreasonable. GPT-4o responds by directly using professional CBT terminology to correct the client’s perspective, which might make the client feel uncomfortable. Llama-3.1-8B also addresses the client’s perspective but does so in a more gentle tone, discussing the underlying reasons for the issue with the client. Llama-3.1-405B delivers the best response. It uses an analogy to help the client view ‘little rewards’ from a different perspective and suggests collaboratively exploring the reasons behind the client’s discomfort. This approach reduces the client’s resistance and fosters greater willingness to engage in self-exploration and self-change.

\section{Conclusion}
In this paper, we propose a new benchmark, \data, to comprehensively evaluate the efficacy of assisting CBT with LLMs. \data contains three levels of tasks ranging from basic CBT knowledge acquisition to therapeutic response generation, totaling five new datasets constructed by domain experts. We experiment with representative LLMs and the results show that models achieve comparable or even better cognitive understanding ability, especially in QA tasks, but still lag behind humans in most generation tasks. We hope our new benchmark provides a meaningful testbed for future research in this area.

\section{Limitations}
In this work, we propose a set of novel tasks aimed at addressing key stages of CBT where LLMs could potentially assist professionals. We collaborated closely with domain experts to select tasks that are both important and challenging, and that are well-suited for AI enhancement. However, annotating datasets in highly specialized fields like mental health is very costly, which has limited the size of our datasets and constrained our ability to scale to additional tasks.
There may be other stages in CBT that could also benefit from AI assistance, which we leave for future research. Additionally, for the level \RN{3} task, we used deliberate practice as an effective proxy to evaluate performance, given the difficulty of acquiring real CBT session data due to privacy concerns. Bridging this gap by approximating real CBT sessions, while respecting ethical and privacy constraints, remains an important direction for future work.

\section{Ethics Statement}
\label{ethics}
\paragraph{IRB (Institutional Review Board) Approval.}
This project is approved by our Institutional Review Board (IRB). 
For all the data annotation and evaluation, we work with our co-authors (clinical psychologists, professors, and social workers) and hire experts from UpWork\footnote{www.upwork.com}. Experts are defined as those with a graduate
degree in clinical psychology, social work, or other related majors and have received at least 5 hours of CBT training. We hired a total of 8 US-based experts, each at an hourly rate of \$60. For some experts, we agreed on a fixed payment based on an estimated completion time, calculated using the same hourly rate.

\paragraph{Informed Consent.}
All experts hired were 18 or older and provided informed consent (Check Appendix~\ref{sec:ann_details} for the consent forms). We did not assess any clinical outcomes. All data collected from the experts were de-identified and consented to be released for research purposes.

\paragraph{System and Data Usages.}
In this work, our primary goal is to systematically evaluate the capabilities of large language models (LLMs) in assisting Cognitive Behavioral Therapy (CBT) in order to provide insights for the future development of AI-based tools to support psychotherapy professionals. All data and systems developed in this work are intended exclusively for academic research purposes. They should not be used as real applications.
We recognize the importance of human expertise and judgment in therapeutic settings and affirm that AI should be used to augment, not substitute, the essential role of trained professionals. Future developments of AI-based systems should not be used directly to interact with real patients without the supervision of professionals. We strongly advocate for rigorous ethical oversight in the deployment of AI-based systems in sensitive areas like mental health (see Appendix~\ref{sec:ethic} for more discussion about ethics considerations). Furthermore, all data used in this study were either publicly available or collected with full transparency and consent, ensuring the privacy and confidentiality of individuals involved. All data collected from the experts were de-identified and consented to be released for research purposes.

\section{Acknowledgment}
We would like to thank Deborah J. Moon and Kate Hardy for providing CBT exam questions and useful discussions.

\bibliography{custom,paperpile}
\appendix

\clearpage
\onecolumn
\section{Label Distribution and Definitions}\label{sec:apx-label}
\begingroup
\setlength{\tabcolsep}{4pt}
\renewcommand{\arraystretch}{3} 
\begin{table*}[h]
    \centering
    \small
    \begin{tabular}{lll}
        \toprule
        \textbf{Belief} & \textbf{\#Label} & \textbf{Definition} \\
        \midrule
        all-or-nothing thinking & 65 & \makecell[l]{Also called black-and-white, polarized, or dichotomous thinking. \\You view a situation in only two categories instead of on a continuum.}  \\
        \cline{3-3}
        mind reading & 47 & \makecell[l]{You believe you know what others are thinking, \\failing to consider other, more likely possibilities.} \\
        \cline{3-3}
        fortune-telling & 44 & \makecell[l]{Also called fortune-telling. You predict the future negatively \\without considering other, more likely outcomes.} \\
        \cline{3-3}
        personalization & 42 & \makecell[l]{You believe others are behaving negatively because of you, \\without considering more plausible explanations for their behavior.} \\
        \cline{3-3}
        emotional reasoning & 36 & \makecell[l]{You think something must be true because you “feel” (actually believe) \\it so strongly, ignoring or discounting evidence to the contrary.}\\
        \cline{3-3}
        overgeneralization & 32 & \makecell[l]{You make a sweeping negative conclusion that \\goes far beyond the current situation.} \\
        \cline{3-3}
        labeling & 29 & \makecell[l]{You put a fixed, global label on yourself or others without considering that \\the evidence  might more reasonably lead to  a less extreme conclusion.} \\
        \cline{3-3}
        should statements & 28 & \makecell[l]{Also called imperatives. You have a precise, fixed idea of how you or others should \\behave, and you overestimate how bad it is that these expectations are not met.} \\
        \cline{3-3}
        magnification & 25 & \makecell[l]{When you evaluate yourself, another person, or a situation, \\you unreasonably magnify the negative and/or minimize the positive.} \\
        \cline{3-3}
        mental filter & 21 & \makecell[l]{Also called selective abstraction. You pay undue attention to \\one negative detail instead of seeing the whole picture.} \\
         \bottomrule
    \end{tabular}
    \label{tab:label_dist_task2}
    \caption{Label distribution and definitions of CBT-CD.}
\end{table*}
\endgroup

\begingroup
\scriptsize
\setlength{\tabcolsep}{4pt}
\renewcommand{\arraystretch}{2} 
\begin{table*}[h]
    \centering
    \small
    \begin{tabular}{lll}
        \toprule
        \textbf{Primary Belief} & \textbf{\#Label} & \textbf{Definition} \\
        \midrule
        helpless & 162 & \makecell[l]{being ineffective—in getting things done, self-protection, \\and/or measuring up to others} \\
        \cline{3-3}
        unlovable & 101 & \makecell[l]{having personal qualities resulting in an inability \\to get or maintain love and intimacy from others} \\
        \cline{3-3}
        worthless & 81 & \makecell[l]{being an immoral sinner or dangerous to others} \\
         \bottomrule
    \end{tabular}
    \label{tab:label_dist_task3}
    \caption{Label distribution and definitions of CBT-PC.}
\end{table*}
\endgroup

\begingroup
\scriptsize
\setlength{\tabcolsep}{4pt}
\renewcommand{\arraystretch}{2} 
\begin{table*}[h]
    \centering
    \caption{Label distribution of CBT-FC.}
    \small
    \begin{tabular}{lll}
        \toprule
        \textbf{Primary Core Belief} & \textbf{Fine-grained Core Belief} & \textbf{\#Label} \\
        \midrule
        \multirow{9}{*}{helpless} & I am incompetent & 16 \\
        & I am helpless & 36 \\
        & I am powerless, weak, vulnerable & 38 \\
        & I am a victim & 29 \\
        & I am needy & 5 \\
        & I am trapped & 28 \\
        & I am out of control & 22 \\
        & I am a failure, loser & 22 \\        
        & I am defective & 31 \\
        \midrule
        \multirow{6}{*}{unlovable} & I am unlovable & 25 \\
        & I am unattractive & 6 \\
        & I am undesirable, unwanted & 27 \\
        & I am bound to be rejected & 29 \\
        & I am bound to be abandoned & 24 \\
        & I am bound to be alone & 21 \\
        \midrule
        \multirow{4}{*}{worthless} & I am worthless, waste & 31 \\
        & I am immoral & 10 \\
        & I am bad - dangerous, toxic, evil & 14 \\
        & I don’t deserve to live & 8 \\
        \bottomrule
    \end{tabular}
    \label{tab:label_dist_task4}
\end{table*}
\endgroup

\begingroup
\scriptsize
\setlength{\tabcolsep}{4pt}
\renewcommand{\arraystretch}{2} 
\begin{table*}[h]
    \centering
    \caption{Statistics of \cbtdp.}
    \small
    \resizebox{1.0\textwidth}{!}{
    \begin{tabular}{lllll}
        \toprule
        \textbf{Category} & \textbf{\#Exercise} & \textbf{\#Beginner} & \textbf{\#Intermediate} & \textbf{\#Advanced} \\
        \midrule
        Explaining the Treatment Rationale for Cognitive Behavioral Therapy & 14 & 4 & 6 & 4 \\
        \midrule
        Establishing Goals & 16 & 5 & 6 & 5 \\
        \midrule
        Negotiating a Session Agenda & 15 & 5 & 5 & 5 \\
        \midrule
        Assigning and Reviewing Between-Session Activities & 15 & 5 & 7 & 5 \\
        \midrule
        Working With Cognitions & 16 & 5 & 5 & 6 \\
        \midrule
        Working With Behaviors & 15 & 5 & 5 & 5 \\
        \midrule
        Working With Emotions & 17 & 5 & 5 & 7 \\
        \midrule
        Adherence Flexibility & 16 & 5 & 6 & 5 \\
        \midrule
        Responding to Therapeutic Alliance Ruptures & 15 & 5 & 5 & 5 \\
        \midrule
        Responding to Client Resistance & 15 & 5 & 5 & 5 \\
        \midrule
        All & 156 & 49 & 55 & 52 \\
        \bottomrule
    \end{tabular}
    }
    \label{tab:label_dist_dp}
\end{table*}
\endgroup

\begingroup
\scriptsize
\setlength{\tabcolsep}{4pt}
\renewcommand{\arraystretch}{2} 
\begin{table*}[h]
    \centering
    \small
    \resizebox{0.98\textwidth}{!}{
    \begin{tabular}{ll}
        \toprule
        \textbf{Category} & \textbf{Criteria} \\
        \midrule
        \makecell[l]{Explaining the Treatment Rationale \\ for Cognitive Behavioral Therapy} & \makecell[l]{Criteria 1: Validate the client's experience, \\
              Criteria 2: Explain the logic of how CBT can be used to address concerns, \\
              Criteria 3: Instill hope for using CBT effectively, \\
              Criteria 4: Set appropriate expectations for the nature and impact of CBT} \\
        \midrule
        Establishing Goals & \makecell[l]{Criteria 1: Suggest CBT-consistent goals and tasks that align with an individualized \\ CBT case formulation, \\
        Criteria 2: Invite the client's input on and agreement with the goals and tasks, \\
        Criteria 3: Demonstrate flexibility, \\
        Criteria 4: Emphasize concrete, actionable, and measurable goals} \\
        \midrule
        Negotiating a Session Agenda & \makecell[l]{Criteria 1: Suggest a CBT-consistent agenda that aligns with the CBT case formulation, \\
        Criteria 2: Invite the client's input on and agreement with the agenda, \\
        Criteria 3: Demonstrate flexibility, \\
        Criteria 4: Maintain some degree of frame or structure (i.e., the agenda can shift, but \\ generally maintain a more or less explicit attempt to delineate expectations for session \\ goals and tasks)
        }\\
        \midrule
        \makecell[l]{Assigning and Reviewing \\ Between-Session Activities} & \makecell[l]{Criteria 1: Provide a basic rationale for the homework assignment, \\
        Criteria 2: Tailor the homework assignment to the client's concerns and needs, \\
        Criteria 3: Ensure that there is adequate agreement on and understanding of the \\ homework assignment, \\
        Criteria 4: Encourage and validate good-faith efforts} \\
        \midrule
        Working With Cognitions & \makecell[l]{Criteria 1: Encourage and facilitate client self-reflection on thoughts and beliefs, \\
        Criteria 2: Maintain an open, exploratory stance, \\
        Criteria 3: Orient the client's attention to thoughts or connections between thoughts \\ and other experiences, \\
        Criteria 4: Emphasize cognitive flexibility, rather than simple thought replacement, \\ and refrain from implying that a particular thought is ``right or wrong"
              } \\
        \midrule
        Working With Behaviors & \makecell[l]{Criteria 1: Maintain a collaborative and curious stance, \\
        Criteria 2: Be clear when suggesting a plan of action and expectations, \\
        Criteria 3: Be positive and supportive of both big and small positive changes, \\
        Criteria 4: Appeal to relevant core learning concepts, including conditioning, \\ reinforcement, the importance of environment, antecedents, and consequences
        } \\
        \midrule
        Working With Emotions & \makecell[l]{Criteria 1: Empathically inquire about clients' emotional experience, \\
        Criteria 2: Actively listen and maintain a supportive tone to clients' emotional \\ disclosures, \\
        Criteria 3: Model tolerance of affect and an approach orientation to clients' strong \\ emotional experience, \\
        Criteria 4: Use psychoeducation to support the importance of clients experiencing \\ versus avoiding their emotion and the maladaptive consequences of negative \\ reinforcement
        } \\
        \midrule
        Adherence Flexibility & \makecell[l]{Criteria 1: Maintain a collaborative and transparent stance, \\
        Criteria 2: Demonstrate empathy, \\
        Criteria 3: Demonstrate openness to explore before determining if a modification is \\ indeed warranted, \\
        Criteria 4: Maintain consistency with a broad CBT orientation
        } \\
        \midrule
        \makecell[l]{Responding to Therapeutic \\ Alliance Ruptures} & \makecell[l]{Criteria 1: Momentarily step away from the CBT change agenda, \\
        Criteria 2: Warmly invite the client to discuss their subjective experience, \\
        Criteria 3: Empathize with the client's thoughts and feelings and invite further \\ disclosure of unhelpful or invalidating occurrences, \\
        Criteria 4: ``Disarm” the client's negative affect toward you or CBT by finding some \\ truth in their disclosures
            } \\
        \midrule
        Responding to Client Resistance & \makecell[l]{Criteria 1: Momentarily step away from the CBT change agenda, \\
        Criteria 2: Empathically explore the client's experience of treatment and you, \\
        Criteria 3: Validate clients' experience and “roll with” versus challenging their \\ resistance, \\
        Criteria 4: Support clients' autonomy and elicit their motivation for pursuing valued \\ directions
            } \\
        \bottomrule
    \end{tabular}
    }
    \caption{Evaluation criteria of \cbtdp.}
    \label{tab:label_dist_dp_cr}
\end{table*}
\endgroup

\clearpage
\section{Examples}\label{sec:apx-example}
\begingroup
\scriptsize
\setlength{\tabcolsep}{4pt}
\begin{table*}[htbp]
\small
\renewcommand{\arraystretch}{3} 
\begin{center}
\resizebox{1.0\textwidth}{!}{
\begin{tabular}{ll}
\toprule
\textbf{Knowledge Types} & \textbf{Example QA Pairs from \cbtqa}\\
\midrule
\multirow{5}{*}{Basic CBT knowledge and concepts} & \makecell[l]{Intermediate beliefs consist of rules, attitudes, and conditional assumptions. Which of the following fits the best as an attitude? \\ \textbf{A. }“Failing is bad”, \textbf{B. }“I must not fail”, \\ \textbf{C: }“If I try, I will fail”, \textbf{D. }“I am a failure”} \\
\cline{2-2}
& \makecell[l]{What type of cognitive error involves shrinking something to make it seem less important? \\ \text{A. }Selective Abstraction, \textbf{B. }Mental filtering, \\ \textbf{C. }Overgeneralization, \textbf{D. }Self-defeated thinking}\\
\cline{2-2}
& \makecell[l]{Which strategy should you employ the most often when you want a client to come to a new understanding? \\ \textbf{A. }Guided discovery, \textbf{B. }Direct advice, \\ \textbf{C: }Indirect advice, \textbf{D. }Disclosing what strategy works the best for you} \\
\cline{2-2}
& \makecell[l]{In the assessment session, you should create a conceptualization/formulation of the client's problem with them. \\When using CBT, which of the following is the most important for a conceptualization/formulation to include? \\ \textbf{A. }How past developmental events impact the problem, \textbf{B. }How thoughts and behaviors relate to the problem, \\ \textbf{C: }How current supports may limit the client's functioning, \textbf{D. }How genetic vulnerabilities result in current problems} \\
\cline{2-2}
& \makecell[l]{CBT helps clients identify links between: \\ \textbf{A. }Health, thoughts, \textbf{B. }Emotions, thoughts, decision-making, \\ \textbf{C: }Thoughts, emotions, behavior, \textbf{D. }Antecedent, behavior, consequence} \\

\midrule
\multirow{5}{*}{Practical CBT knowledge} & \makecell[l]{How might a social worker use the principles of CBT approach when working with a client who is experiencing suicidal thoughts? \\ \textbf{A. }By encouraging the client to suppress their thoughts and focus on positive affirmations, \\ \textbf{B. }By providing temporary shelter and financial assistance to alleviate immediate stressors, \\ \textbf{C. }By collaboratively identifying and challenging distorted thought patterns contributing to suicidal thoughts, \\
\textbf{D. } No options align with any practices considered to be CBT} \\
\cline{2-2}
& \makecell[l]{A primary concern during the final stages of a cognitive behavioral group is which of the following? \\ \textbf{A. }promoting transfer of learning, \textbf{B. }developing a therapeutic contract, \\ \textbf{C. }role-playing various situations \textbf{D. } establishing baseline data} \\
\cline{2-2}
& \makecell[l]{Early in treatment you advise clients that the way to get better is? \\ \textbf{A. }By making small changes in their thinking and behaviors each day, \textbf{B. }By making large changes to their core beliefs, \\ \textbf{C. }By gaining insight regarding their developmental history \textbf{D. } By changing their emotions} \\
\cline{2-2}
& \makecell[l]{When helping clients evaluate automatic thoughts, therapists should generally help clients evaluate which aspects of those thoughts? \\ \textbf{A. }accuracy and/or intensity, \textbf{B. }intensity and/or utility, \\ \textbf{C. }accuracy and/or utility} \\
\cline{2-2}
& \makecell[l]{Which of the following typically improve depressive symptoms (select all that apply)? \\ \textbf{A. }Interpersonal interactions, \textbf{B. }Activities that have the potential for pleasure, \\ \textbf{C. }Activities that have the potential for a sense of mastery, \textbf{D.} Repeatedly thinking about the reasons one is depressed} \\
\midrule

\multirow{5}{*}{Case studies} & \makecell[l]{The client says “I feel like he doesn’t care about me”. “I feel like he doesn’t care about me” is a? \\ \textbf{A. }Thought, \textbf{B. }Emotion, \\ \textbf{C. }Physiological reaction/sensation, \textbf{D. }Feeling} \\
\cline{2-2}
& \makecell[l]{A manager tells you (the therapist) that the new employee, who has been doing an excellent job, requested a meeting with him. \\The manager anticipates that the new employee plans to resign. Which is the best description of the manager's cognitive error? \\ \textbf{A. }Thinking from an all-or-nothing perspective, \textbf{B. }Always thinking the worst will occur without considering positive outcomes, \\ \textbf{C. }Viewing only selected negative evidence while editing out positive aspects, \textbf{D. }Undervaluing the positive significance of an event} \\
\cline{2-2}
& \makecell[l]{The client’s mood has been improving since they’ve been attending treatment. They could attribute this improvement to many different things. \\ Which of the below do you want them to attribute it to the most? \\ \textbf{A. }The positive changes they have been making in their thinking and behaviors, \textbf{B. }The support their therapist has given them, \\ \textbf{C. }Events in their environment that have changed, \textbf{D. }The randomness of their mood} \\
\cline{2-2}
& \makecell[l]{Ms. T. has a number of long-standing negative attitudes such as "I'll never succeed" and "I have to be perfect to be accepted." \\No matter how hard she tries she always seems to think that she is "not measuring up." \\What cognitive therapy procedures do you think might help her? \\ \textbf{A. }Listing advantages and disadvantages, \textbf{B. }Breaking out, \\ \textbf{C. }Listing schemas, \textbf{D. }Thought recording} \\
\cline{2-2}
& \makecell[l]{A client says, "I don't feel like I'm doing a good job at work" \\What question can you ask next to start the downward arrow technique to identify their core belief? \\ \textbf{A. }"How did your parents give feedback when you were younger?", \textbf{B. }"That's not true, you're a great worker!", \\ \textbf{C. }"What are you worried that might mean?", \textbf{D. }"What is the evidence for and against this thought?"} \\

\midrule
\multirow{5}{*}{Others} & \makecell[l]{Among the following principles, which one does not align with the core values of social work practice? \\ \textbf{A. }Respect for the inherent dignity and worth of every individual,  \\ \textbf{B. }Promotion of client self-determination and autonomy, \\ \textbf{C. }Maintenance of strict neutrality and avoidance of taking sides, \\ \textbf{D. }Commitment to social justice and advocacy for marginalized populations} \\
\cline{2-2}
& \makecell[l]{Which of the following best describes the collaborative nature of a client and a therapist in social work practice? \\ \textbf{A. }The therapist provides solutions to the client's problems without considering the client's perspective,  \\ \textbf{B. }The therapist and client work together as partners to identify goals and develop strategies for change, \\ \textbf{C. }The therapist imposes their own values and beliefs onto the client without discussion, \\ \textbf{D. }The therapist directs the client's treatment without input or collaboration from the client} \\
\cline{2-2}
& \makecell[l]{In the field of social work, what is the primary goal of intervention? \\ \textbf{A. }Providing financial assistance to individuals in need,  \\ \textbf{B. }Offering temporary shelter to homeless populations, \\ \textbf{C. }Enhancing the well-being and social functioning of individuals, families, and communities, \\ \textbf{D. }Administering psychological assessments to diagnose mental health disorders} \\
\cline{2-2}
& \makecell[l]{A social worker needs to address the specific needs of oppressed populations, such as marginalized racial and ethnic groups. \\What would be the best approach through a CBT framework? \\ \textbf{A. }Allowing others in the community to take part and take the lead. Mainly because as social workers we should allow others to take charge,  \\ \textbf{B. }By encouraging clients to conform to dominant cultural norms to reduce societal backlash, \\ \textbf{C. }By validating that such issues and problems exist, and then seeing if your agency has any funding to help purchase food for families impacted, \\ \textbf{D. }By validating clients' experiences of discrimination and integrating culturally sensitive interventions} \\
\cline{2-2}
& \makecell[l]{How might therapists incorporate LGBTQ-affirmative approaches to address the unique challenges faced by LGBTQ individuals? \\ \textbf{A. }By avoiding discussions about clients' LGBTQ identities to focus solely on symptom management, \\ \textbf{B. }By validating the identities of LGBTQ-identified individuals through demonstrating acceptance and affirming their experiences, \\ \textbf{C. }By challenging clients' LGBTQ identity as a potential cause of their mental health struggles, \\ \textbf{D. }By encouraging clients to suppress their LGBTQ identity to avoid social stigma} \\
\bottomrule
\end{tabular}
}
\label{tab:example-task1}
\caption{Example QA pairs of \cbtqa.}
\end{center}
\end{table*}
\endgroup

\begin{figure*}[h]
\centering
\includegraphics[width=\linewidth]{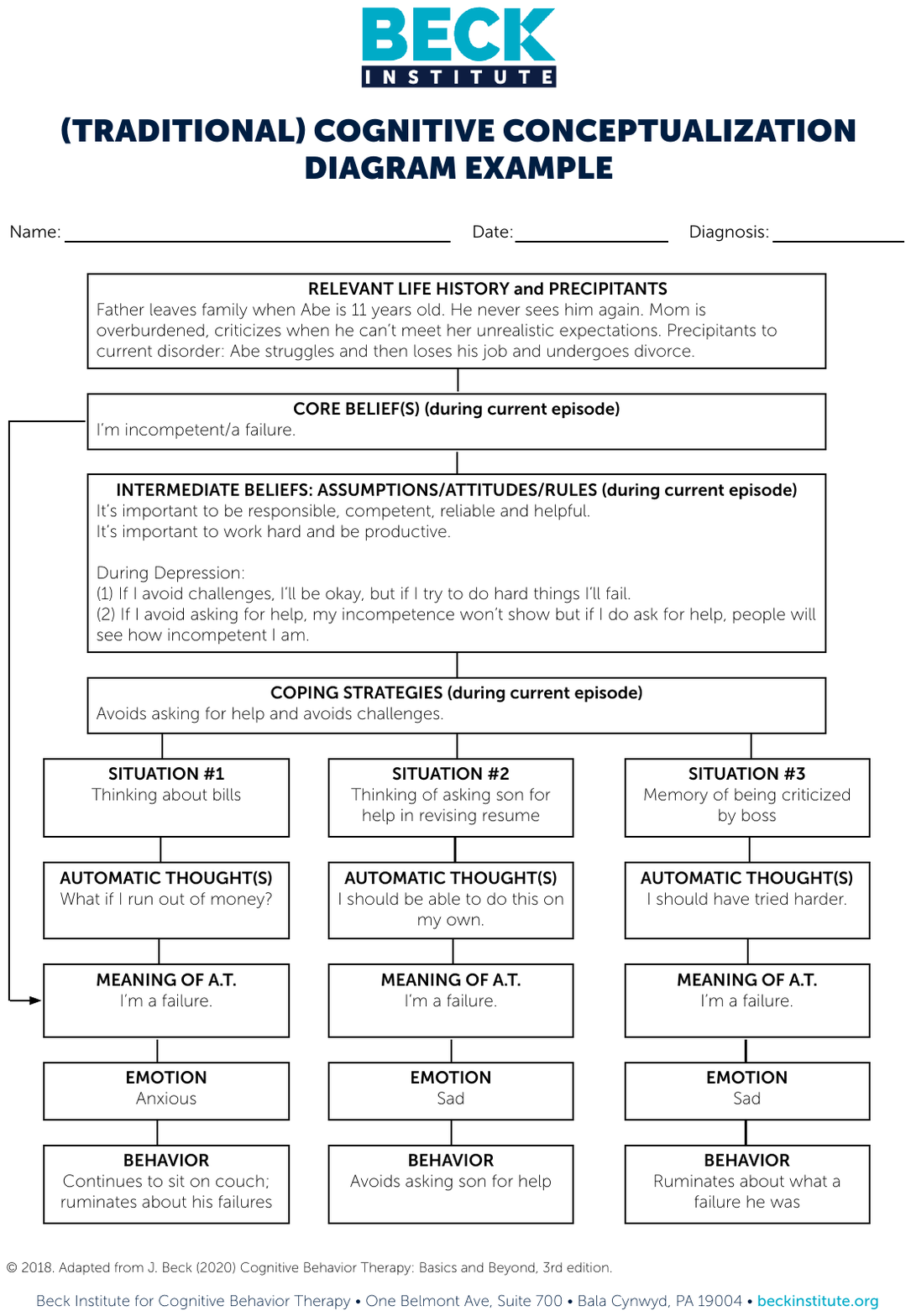}
\caption{An example cognitive model from \citet{beck2020cognitive}.}
\label{fig:cog_model}
\end{figure*} 

\begingroup
\setlength{\tabcolsep}{4pt}
\begin{table*}[t]
\scriptsize
\renewcommand{\arraystretch}{3} 
\begin{center}
\resizebox{0.96\textwidth}{!}{
\begin{tabular}{p{8cm}p{4cm}}
\toprule
Inputs & Labels \\
\midrule
\textbf{Situation}: I have been with my girlfriend for 5 years and we have a history of a strong sexual intimacy connection. This has been getting worse over time. \textbf{Thoughts}: It is my girlfriend's sex drive that is the problem. & overgeneralization; labeling \\

\textbf{Situation}: I am struggling at the moment and yet I am aware that I should be feeling good. \textbf{Thoughts}: What I am feeling and going through are unacceptable and wrong. There is a right and wrong way to go through this and I am doing it the wrong way. & should statements; emotional reasoning \\

\textbf{Situation}: The individual has recently discovered that their mom has cheated on their dad four times in the past. Each time, the parents attended counseling, where the mom initially denied her infidelity but eventually admitted to it, blaming their dad for not meeting her needs. \textbf{Thoughts}: I cannot trust anyone, why is everyone a liar. My mom has betrayed me. Of course something like this is happening to me, it only happens to me. & all-or-nothing thinking; overgeneralization; emotional reasoning \\

\textbf{Situation}: I experienced a recent setback with my OCD symptoms that had been in remission \textbf{Thoughts}: I have no longer won the battle against my OCD & all-or-nothing thinking; magnification \\

\textbf{Situation}: My anxiety used to be worse a couple of years ago, but now im just paranoid. Im always thinking that people think im a loser, and I wont leave the house without makeup, because i am horrified about what other people think about me. \textbf{Thoughts}: Because my anxiety felt worse before, this must be a normal level of anxiety and paranoia, so my conclusions make more sense. People do think I’m a loser or that I’m ugly. I at least wear makeup so they can’t see how ugly I am, and if I avoid people altogether, then they won’t know how much of a loser I am or how boring I am. I don’t like who I am, and whatever people think of me is probably way worse than I think it is. & overgeneralization; emotional reasoning; mind reading \\
\bottomrule
\end{tabular}
}
\end{center}
\caption{Examples from \cbtcd (Cognitive Distortion Classification).}
\end{table*}
\endgroup

\begingroup
\setlength{\tabcolsep}{4pt}
\begin{table*}[t]
\scriptsize
\renewcommand{\arraystretch}{3} 
\begin{center}
\resizebox{0.96\textwidth}{!}{
\begin{tabular}{p{8cm}p{3cm}}
\toprule
Inputs & Labels \\
\midrule
\textbf{Situation}: I've been dating this guy for 2 years, and when things are good, they're great. We have deep conversations about various topics, share laughs, and seem to have a strong connection. However, every few months, he suddenly becomes distant. The usual daily communication slows down, his tone becomes cold, and there's a noticeable disconnect. After giving it some space and bringing it up gently, he admits to withdrawing and usually provides a reason for his behavior, such as feeling disrespected or abandoned. We address the issues, talk about feelings and needs, and find better ways to communicate. He has a therapist, and I encourage him to work on these issues, but the cycle repeats. \textbf{Thoughts}: I am not enough for him.
Why do I always get hurt like this. I always get stuck in these endless cycles, and I am unlovable. & helpless; unlovable \\

\textbf{Situation}: I would move out, but I don’t have the money for that nor do I have the job. Whenever I get I job, I take too many sick days out because of the bruises and scarring on my face are hard to hide. \textbf{Thoughts}: I can't move out until I have a better financial situation. I must remain in this situation until I have more money. The only way out is to be more financially secure. I cannot let other people know or help me. I have to sick days to hide the bruising on my face. Other people can't find out what's happening. I must solve this all on my own. & helpless; unlovable \\

\textbf{Situation}: Ever since, I have known in the back of my mind that he still does this, and have gotten up in the middle of the night a few times to find him masturbating and video chatting publicly or with females who are at least scantily clad, if not nude and masturbating themselves. When I confront him about it, he says he has a public sex fetish and needs to fulfill it, believing that it helps keep him „faithful‰ to me since I am no longer interested. \textbf{Thoughts}: He calls that faithful? It doesn’t feel faithful. But I don’t know if I even have the right to feel the way I do about it – after all, that’s pretty much how our relationship started. Am I even allowed to be upset? I’m the one that changed, so it feels like it’s my fault. I shouldn’t be blaming him. He tried to get me involved, but when I said I wasn’t into it, he didn’t push, he just continued on doing what made him happy. Plus, he’s not doing this to hurt me, so I shouldn’t feel hurt by it. I should be happy for him that he is feeling fulfilled. But I feel sad, like he doesn’t want me anymore. & helpless; unlovable; worthless \\

\textbf{Situation}: Do I have schizophrenia or something \textbf{Thoughts}: I am scared and don't know what is happening. I fear this could be something very serious. There is something wrong with me. I don't know what to do. & helpless \\

\textbf{Situation}: I was born with Attention Deficit Hyperactivity Disorder, O.C.D. and I am also transgendered. As a result, I was a very difficult child to raise. There was constant fighting between myself and my parents. At the age of ten, I was sent away to an institution for two years. While there, I was physically, emotionally, and sexually abused. I couldn’t talk to anyone because I had no trust in adults. \textbf{Thoughts}: I was born defective and wrong, so that’s why my parents had such a hard time raising me. I was hard to love because I was so different. It was my fault that they had to send me away; I needed to be fixed. Then, I was abused while I was away but I couldn’t talk to anyone about it because I was scared that if I said anything, I’d be sent even further away. & helpless; unlovable; worthless \\
\bottomrule
\end{tabular}
}
\end{center}
\caption{Examples from \cbtmc (Primary Core Belief Classification).}
\end{table*}
\endgroup

\begingroup
\setlength{\tabcolsep}{4pt}
\begin{table*}[t]
\scriptsize
\renewcommand{\arraystretch}{3} 
\begin{center}
\resizebox{0.96\textwidth}{!}{
\begin{tabular}{p{8cm}p{3cm}}
\toprule
Inputs & Labels \\
\midrule
\textbf{Situation}: My relationship has always been in trouble because of my wife’s suspicious nature. \textbf{Thoughts}: - This is all my wife's fault, not mine. If she stopped being suspicious, then all of our marital problems would be solved. There is nothing I can do about this; she is the one who needs to change. & I am incompetent; I am powerless, weak, vulnerable; I am a victim \\

\textbf{Situation}: My relationship with my mom has deteriorated due to a lie I told back in February. I took full responsibility for it, but I'm aware that my actions, such as hiding my depression from her and seeing a psychologist without her knowledge, were not acceptable. I felt that she wouldn't understand because she's always belittled my emotions, accusing me of exaggerating and seeking attention whenever I express myself emotionally. This led me to feel ashamed and afraid of facing her criticism, so I chose to keep my struggles hidden. \textbf{Thoughts}: I'm a failure for lying and hiding things from my mom.
She'll never understand or accept me for who I am. I'm just seeking attention by expressing my feelings. & I am trapped; I am a failure, loser; I am unlovable; I am undesirable, unwanted; I am bound to be rejected; I am worthless, waste \\

\textbf{Situation}: ...although we all had a lovely day I noticed he was distant... \textbf{Thoughts}: I am confused by this situation, I do not know how to interpret this or respond. I cannot trust myself. I cannot trust my own thoughts about this. & I am incompetent; I am powerless, weak, vulnerable \\

\textbf{Situation}: I have been backwards... I was forced... \textbf{Thoughts}: My sleep habits are abnormal. I should feel conscientious and weird about my sleep patterns. I may be judged for being different. I regret or resent having been forced. & I am defective \\

\textbf{Situation}: I never told anyone the truth about it or should I say I never admitted it to anyone... I swore I would never cheat on my husband because I never forgave myself for cheating on him with the married guy... I hated myself for that. \textbf{Thoughts}: I feel guilt for what I did. I blame myself for what happened. I am a bad person. I regret what I did. I wanted to make amends for what I did. I am only worthy of love if I hide my past mistakes. I feel shame for what I did. & I am bound to be rejected; I am bound to be abandoned; I am immoral \\
\bottomrule
\end{tabular}
}
\end{center}
\caption{Examples from \cbtfc (Fine-grained Core Belief Classification).}
\end{table*}
\endgroup

\clearpage
\section{Task Prompts}\label{sec:task-prompts}
\begingroup
\begin{table*}[h]
    \centering
    \small
    \resizebox{1.0\textwidth}{!}{
    \begin{tabular}{p{\textwidth}}
        \toprule
        You are taking a CBT exam and doing multiple-choice questions. Each question has only one right choice. \\
        Which core principle underlies Cognitive Behavioral Therapy? \\
        a: Acceptance and Commitment \\
        b: Mindfulness \\
        c: The influence of desires \\
        d: The interconnection between thoughts, feelings, and behaviors \\
        Answer: \\
        Please only output the letter corresponding to the choice. \\
        \bottomrule
    \end{tabular}
    }
    \caption{Prompt example for CBT Exam QA.}
    \label{tab:apx-prompt-task1}
\end{table*}
\endgroup
\begingroup
\begin{table*}[h]
    \centering
    \small
    \resizebox{1.0\textwidth}{!}{
    \begin{tabular}{p{\linewidth}}
        \toprule
        You are a CBT therapist and now need to determine the cognitive distortions of a patient from his current situation and thoughts. Each patient may have **up to 3** distortions. \\
        Situation: I’m depressed but nobody knows it. I do not have any friends. This started at age 11. \\
        Thoughts: I cannot make friends and have no one to hang out with. Therefore, I am always going to be alone and depressed. \\
        Question: what distortions this patient has? \\
        Choices: \\
        a: all-or-nothing thinking \\
        b: overgeneralization \\
        c: mental filter \\
        d: should statements \\
        e: labeling \\
        f: personalization \\
        g: magnification \\
        h: emotional reasoning \\
        i: mind reading \\
        j: fortune-telling \\
        Answer: \\
        Please only output the letters corresponding to the choices. Multiple choices should be separated by a comma. \\
        \bottomrule
    \end{tabular}
    }
    \caption{Prompt example for Cognitive Distortion Classification.}
    \label{tab:apx-prompt-task2}
\end{table*}
\endgroup
\begingroup
\begin{table*}[h]
    \centering
    \small
    \resizebox{1.0\textwidth}{!}{
    \begin{tabular}{p{\linewidth}}
        \toprule
        You are a CBT therapist and now need to determine the major core beliefs of a patient from his current situation and thoughts. Each patient may have multiple core beliefs. \\
        Situation: I have a history of being hurt in relationships so decided to take a break from dating. I am now in a relationship but keep hurting him and taking everything out on him. \\
        Thoughts: I am hurting my boyfriend because I have been hurt in the past. I have so many problems. \\
        Question: what core beliefs this patient has? \\
        a: helpless \\
        b: unlovable \\
        c: worthless \\
        Answer: \\
        Please only output the letters corresponding to the choices. Multiple choices should be separated by a comma. \\
        \bottomrule
    \end{tabular}
    }
    \caption{Prompt example for Primary Core Belief Classification.}
    \label{tab:apx-prompt-task3}
\end{table*}
\endgroup
\begingroup
\begin{table*}[h]
    \centering
    \small
    \resizebox{1.0\textwidth}{!}{
    \begin{tabular}{p{\linewidth}}
        \toprule
        You are a CBT therapist and now need to determine the fine-grained beliefs of a patient from his current situation and thoughts. Each patient may have **up to 9** fine-grained beliefs. Now answer the following question: \\
        Situation: The individual has recently discovered that their mom has cheated on their dad four times in the past. Each time, the parents attended counseling, where the mom initially denied her infidelity but eventually admitted to it, blaming their dad for not meeting her needs. \\
        Thoughts: I cannot trust anyone, why is everyone a liar. My mom has betrayed me. Of course something like this is happening to me, it only happens to me. \\
        Question: what fine-grained beliefs this patient has? \\
        Choices: \\
        a: I am incompetent \\
        b: I am helpless \\
        c: I am powerless, weak, vulnerable \\
        d: I am a victim \\
        e: I am needy \\
        f: I am trapped \\
        g: I am out of control \\
        h: I am a failure, loser \\
        i: I am defective \\
        j: I am unlovable \\
        k: I am unattractive \\
        l: I am undesirable, unwanted \\
        m: I am bound to be rejected \\
        n: I am bound to be abandoned \\
        o: I am bound to be alone \\
        p: I am worthless, waste \\
        q: I am immoral \\
        r: I am bad - dangerous, toxic, evil \\
        s: I don’t deserve to live \\
        Answer: \\
        Please only output the letters corresponding to the choices. Multiple choices should be separated by a comma. \\
        \bottomrule
    \end{tabular}
    }
    \caption{Prompt example for Fine-grained Core Belief Classification.}
    \label{tab:apx-prompt-task4}
\end{table*}
\endgroup

\begingroup
\begin{table*}[h]
    \centering
    \small
    \resizebox{1.0\textwidth}{!}{
    \begin{tabular}{p{\linewidth}}
    \toprule
    \multicolumn{1}{p{\linewidth}}{Now, you are a professional therapist using cognitive-behavioral therapy (CBT) in a session with a client. Answer the given client statement in around "\{gold response length\}" words. At the beginning of the client statement, we have the client's emotion in brackets, representing the client's emotional status when speaking the statement. You do not need to include emotions in brackets in your response.} \\
    \bottomrule
    \end{tabular}
    }
    \caption{Prompt example for Therapeutic Response Generation.}
    \label{tab:apx-prompt-task5}
\end{table*}
\endgroup

\clearpage
\section{Detailed Model Performance}\label{sec:apx-class-metrics}
\begin{figure*}[htbp]  
    \centering  
    
    \begin{subfigure}[b]{0.45\textwidth}  
        \centering
        \includegraphics[width=\textwidth]{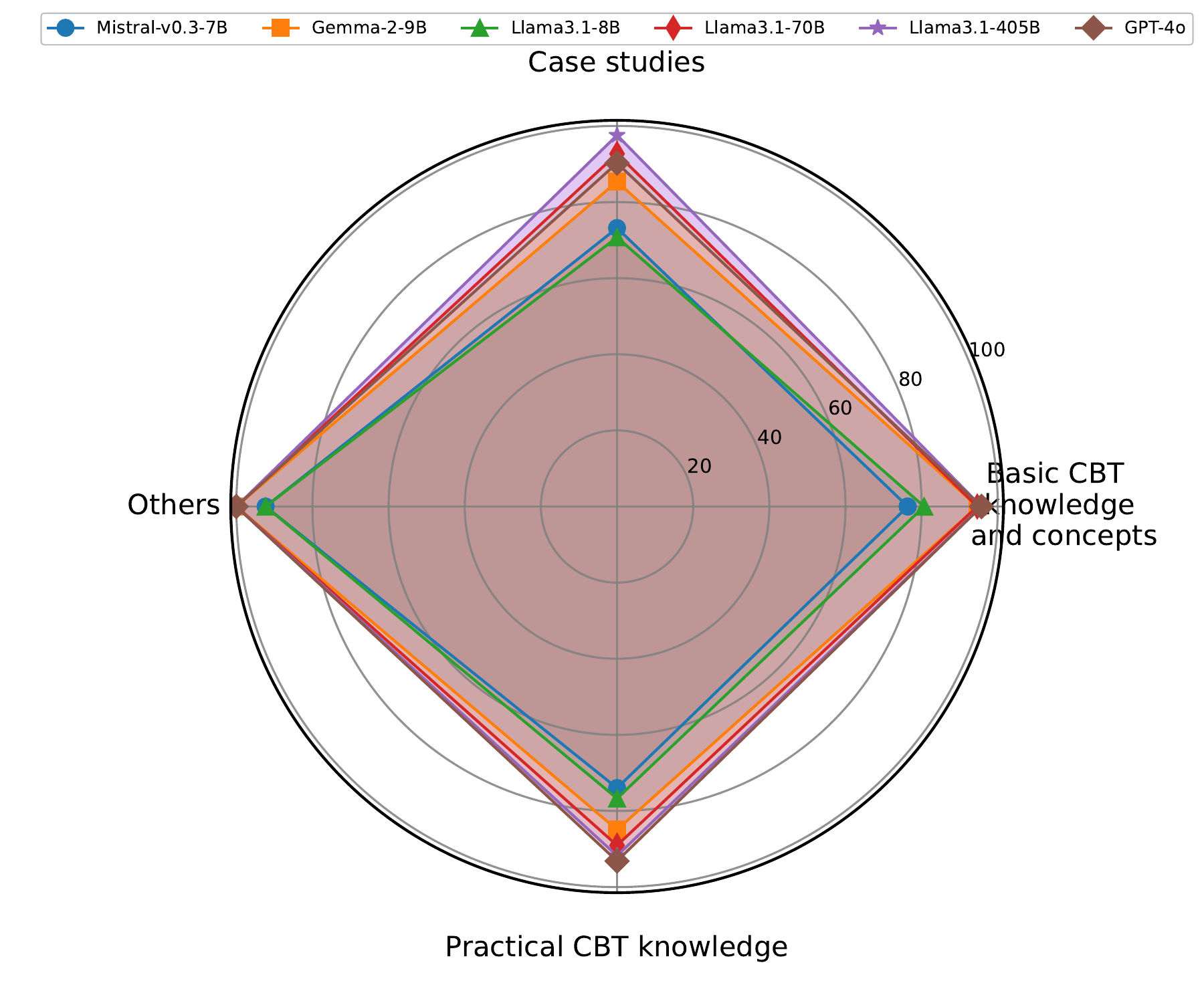}  
        \caption{\cbtqa}
        \label{subfig:task1}
    \end{subfigure}
    \hfill  
    \begin{subfigure}[b]{0.45\textwidth}
        \centering
        \includegraphics[width=\textwidth]{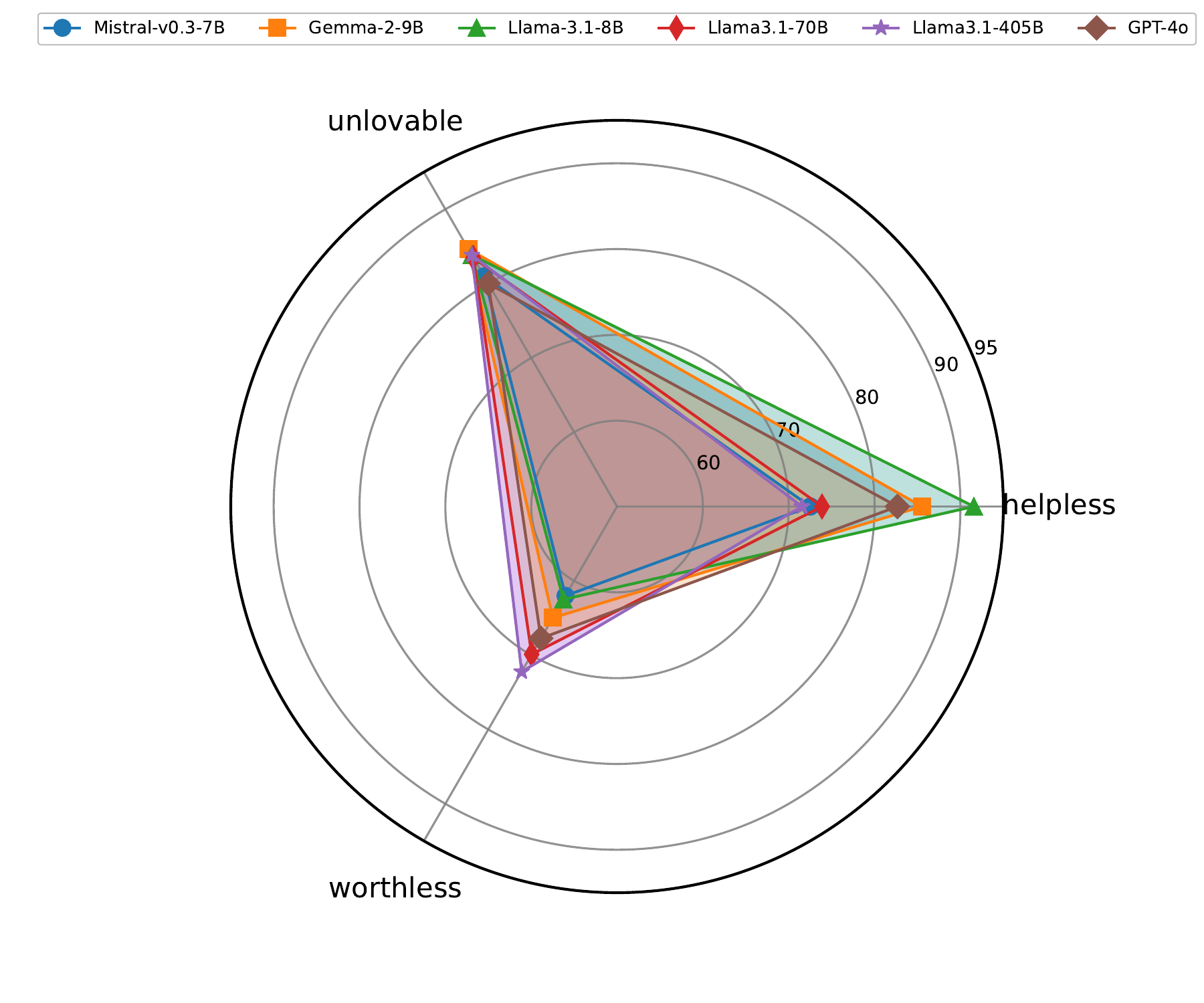}
        \caption{\cbtmc}
        \label{subfig:task3}
    \end{subfigure}
    \caption{Detailed accuracies on different types of knowledge for \cbtqa and the F1 score of each label for \cbtmc.}
\end{figure*}\label{detailed}

\clearpage
\section{Performance Breakdown by Criteria and Exercise}\label{sec:apx-criteria-breakdown}
Here we show the breakdown of model performance under 4 criteria on 10 exercises in Table \ref{exp3: criteria_class}. It is noticeable that all models achieve negative results under four criteria on questions from exercise 10.

\begin{table*}[h!]
\centering
\begin{tabular}{ccrrr}
\toprule
Exercise & Criteria  & Llama-3.1-405B vs. ref & Llama-3.1-8B vs. ref & GPT-4o vs. ref \\ 
\midrule
\multirow{4}{*}{Exercise 1} & Criteria 1 & 0.00 & -0.36 & -0.64 \\
                         & Criteria 2 & 0.64 & 0.79 & 0.93 \\ 
                         & Criteria 3 & 0.36 & 0.36 & 0.43 \\  
                         & Criteria 4 & 0.50 & 0.43 & 0.29 \\ \hline
\multirow{4}{*}{Exercise 2} & Criteria 1 & 0.34 & -0.25 & 0.25 \\  
                         & Criteria 2 & 0.06 & 0.13 & -0.25 \\  
                         & Criteria 3 & -0.38 & -0.13 & -0.63 \\  
                         & Criteria 4 & 0.56 & 0.13 & 0.81 \\ \hline
\multirow{4}{*}{Exercise 3} & Criteria 1 & 0.00 & -0.40 & -0.53 \\
                         & Criteria 2 & 0.07 & -0.27 & -0.27 \\  
                         & Criteria 3 & 0.07 & -0.07 & 0.13 \\  
                         & Criteria 4 & -0.33 & -0.33 & -0.20 \\ \hline
\multirow{4}{*}{Exercise 4} & Criteria 1 & -0.06 & -0.53 & 0.24 \\
                         & Criteria 2 & -0.29 & -0.47 & 0.12 \\  
                         & Criteria 3 & -0.18 & -0.18 & -0.41 \\  
                         & Criteria 4 & -0.12 & -0.47 & 0.00 \\ \hline
\multirow{4}{*}{Exercise 5} & Criteria 1 & -0.06 & 0.00 & -0.44 \\  
                         & Criteria 2 & -0.31 & -0.44 & -0.50 \\  
                         & Criteria 3 & 0.25 & 0.63 & 0.00 \\  
                         & Criteria 4 & 0.06 & -0.19 & 0.06 \\ \hline
\multirow{4}{*}{Exercise 6} & Criteria 1 & 0.13 & 0.33 & -0.40 \\   
                         & Criteria 2 & 0.20 & -0.20 & 0.13 \\  
                         & Criteria 3 & -0.07 & 0.00 & 0.13 \\  
                         & Criteria 4 & -0.13 & -0.33 & -0.07 \\ \hline
\multirow{4}{*}{Exercise 7} & Criteria 1 & 0.41 & 0.18 & 0.06 \\  
                         & Criteria 2 & -0.12 & 0.00 & -0.18 \\  
                         & Criteria 3 & 0.00 & -0.35 & -0.29 \\  
                         & Criteria 4 & -0.06 & 0.12 & 0.24 \\ \hline
\multirow{4}{*}{Exercise 8} & Criteria 1 & -0.06 & 0.06 & -0.25 \\  
                         & Criteria 2 & -0.56 & 0.38 & -0.06 \\  
                         & Criteria 3 & 0.06 & 0.38 & -0.25 \\  
                         & Criteria 4 & 0.25 & 0.13 & 0.00 \\ \hline
\multirow{4}{*}{Exercise 9} & Criteria 1 & 0.07 & -0.53 & -0.40 \\  
                         & Criteria 2 & 0.07 & 0.13 & -0.13 \\  
                         & Criteria 3 & -0.27 & -0.13 & -0.53 \\  
                         & Criteria 4 & -0.60 & -0.73 & -0.33 \\ \hline
\multirow{4}{*}{Exercise 10} & Criteria 1 & -0.27 & -0.53 & -0.60 \\  
                         & Criteria 2 & -0.20 & -0.33 & -0.60 \\  
                         & Criteria 3 & -0.07 & -0.47 & -0.47 \\  
                         & Criteria 4 & -0.07 & -0.47 & -0.33 \\ 
\bottomrule
\end{tabular}
\caption{All model vs. ref (reference) results under 4 criteria on 10 exercises.}\label{exp3: criteria_class}
\end{table*}

\begin{table*}[ht]
\centering
\small
\resizebox{1.0\textwidth}{!}{
\begin{tabular}{p{3cm}p{13cm}}
\toprule
Client Statement & [Agitated] Honestly, my main problem is my partner.\\
\midrule
{Llama-3.1-8B} & Let's break that down. It sounds like you're feeling frustrated with your partner. Can you tell me more about what specifically is causing you distress in your relationship? What are some specific issues or behaviors that you'd like to work on or change?	 \\
\midrule
{Human Reference} & I’d like to hear more details about that. If relationship issues are most concerning to you, we can focus our work on that area. In our individual work together, we’ll need to focus on things that you can change. Part of our session today can be brainstorming what some of those things might be. \\
\bottomrule
\end{tabular}
}
\caption{Exercise 2 example: comparison between model and reference.}\label{case_study: creteria}
\end{table*}

\begin{table*}[ht]
\centering
\small
\resizebox{1.0\textwidth}{!}{
\begin{tabular}{p{3cm}|p{10cm}|p{3cm}}
\toprule
Client Statement & [Frustrated] I don’t like it when you ask me if I have “catastrophic thoughts.” Aren’t catastrophes reserved for wars and natural disasters? So, no, I don’t think I catastrophize. Please don’t ask again. &  Score\\
\hline
{Llama-3.1-8B} & I understand your frustration, and it's valid to feel that way when we challenge your thoughts. However, catastrophizing isn't limited to wars and natural disasters; it's a thought pattern where we exaggerate the consequences of a situation, making it seem more severe or overwhelming than it actually is. Can we explore what makes you uncomfortable about this concept, and how it relates to your current concerns?	& - \\
\hline
{Human Reference} & It seems like my words upset you, and I totally get it. Here I am using a CBT term that I assumed would be relevant you, when in fact it’s not. I really should not have made that assumption, and I’ll do better to check in. In fact, I wonder if you find any part of focusing on your cognitions to be useful? & - \\
\hline
Criteria 1 & Momentarily step away from the CBT change agenda   &  -1 \\  \hline
Criteria 2 &  Empathically explore the client's experience of treatment and you	  &  -1  \\ \hline
Criteria 3 &  Validate clients' experience and “roll with” versus challenging their resistance  &  0  \\ \hline
Criteria 4 &  Support clients' autonomy and elicit their motivation for pursuing valued directions  &  -1 \\
\bottomrule
\end{tabular}
}
\caption{Exercise 10 example: comparison between model and reference.}\label{case_study: e2_1}
\end{table*}

\begin{table*}[ht]
\centering
\small
\resizebox{1.0\textwidth}{!}{
\begin{tabular}{p{3cm}|p{10cm}|p{3cm}}
\toprule
Client Statement & [Defeated] You know, I’ve felt a lot better these past few weeks, and I honestly wonder if it’s because I stopped doing those exercises you gave me. I mean, maybe writing down my thoughts and trying to relax is making me worse . . . maybe this is just who I am, and this is the best that I can feel, and I just need to learn to live with that. &  Score\\
\hline
{GPT-4o} &  It sounds like you're feeling quite discouraged right now. It's great that you’ve been feeling better, but it’s also important to consider if avoiding the exercises might have short-term relief but long-term consequences. Sometimes, initial resistance to new techniques is a part of the process. Could we explore if there might be aspects of the exercises that could be adapted to better fit your needs? CBT works best with some trial and error, and finding what genuinely helps you long-term is our goal.	& - \\
\hline
{Human Reference} & I can hear in your voice how painful it is to feel like this might be the best you can ever feel. And, it sounds like one of the major reasons you’re feeling this way is that the exercises I’ve given you may be making things worse instead of better. So, I want to say that different strategies work for different people, and there are all kinds of strategies that we can try to help you get to where you would like to be. Can we discuss what that might look for you?	
 & - \\
\hline
Criteria 1 & Momentarily step away from the CBT change agenda   &  0 \\  \hline
Criteria 2 &  Empathically explore the client's experience of treatment and you		  &  -1  \\ \hline
Criteria 3 &  Validate clients' experience and “roll with” versus challenging their resistance &  -1  \\ \hline
Criteria 4 &  Support clients' autonomy and elicit their motivation for pursuing valued directions &  -1 \\
\bottomrule
\end{tabular}
}
\caption{Exercise 10 example: comparison between model and reference.}\label{case_study: e2_2}
\end{table*}

\clearpage
\section{Win-tie-loss Analysis}\label{sec:win-tie}

\begin{figure*}[h]
\centering
\includegraphics[width=0.98\linewidth]{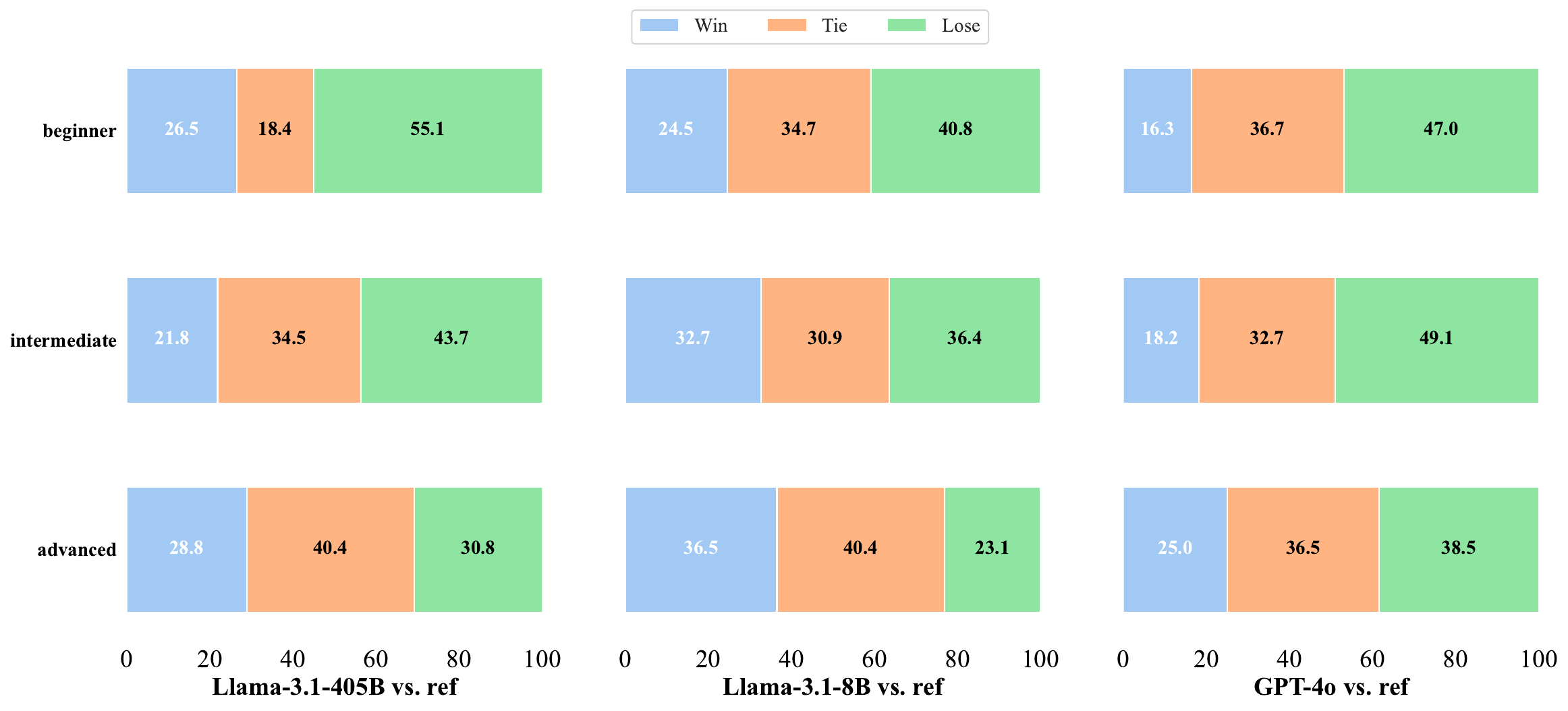}
\caption{The win-tie-loss comparison among different models on three difficulty levels.}\label{fig: win_tie}
\end{figure*}

In this section, we additionally provide the win-tie-loss results by difficulty level in Figure \ref{fig: win_tie}. For win-tie-loss calculation, we consider results with positive scores such as scores 1 and 2 as win, negative scores such as -1, and -2 as loss, and score 0 as tie.
It is clear that the model responses seldom win but usually tie with the reference, and the reference wins more often. This confirms that although the models can already provide useful CBT assistance compared with human experts, but still lag and need further improvements.

\clearpage
\section{Error Cases}\label{sec:error-ans-level-12}
\begingroup
\small
\setlength{\tabcolsep}{4pt}
\begin{table*}[t]

    \centering
    \small
    \begin{tabular}{p{8cm}|l|p{4cm}}
        \toprule
        \textbf{Case} & \textbf{\makecell[l]{Unrecognized \\Disorder}} & \textbf{Reason} \\
        \midrule
        \textbf{Situation}: My anxiety used to be worse a couple of years ago, but now I'm just paranoid. I'm always thinking that people think I'm a loser, and I won't leave the house without makeup, because I am horrified about what other people think about me. \textbf{Thoughts}: Because my anxiety felt worse before, this must be a normal level of anxiety and paranoia, so my conclusions make more sense. People do think I'm a loser or that I'm ugly. I at least wear makeup so they can't see how ugly I am, and if I avoid people altogether, then they won't know how much of a loser I am or how boring I am. I don't like who I am, and whatever people think of me is probably way worse than I think it is. & Overgeneralization &  The model focuses on the client’s fear of judgment and anxiety in social situations rather than recognizing the generalized, negative self-beliefs indicated by "People do think I'm a loser or that I'm ugly". \\
        \midrule
        \textbf{Situation}: It's all my fault most likely all I do is constantly worry about everything. \textbf{Thoughts}: I should be able to control others around me, and when I can't, I cope with it by worrying. If someone is making decisions I disagree with, I should worry. There is something wrong with me that I can't control others. I am doing everything wrong. & Should statements &  The model fail to focus on the client's statement "I should be able to control others around me", which is a clear indicator of "Should statements" disorder.  \\
        \midrule
        \textbf{Situation}: I have been recently dealing with weird behavior...I have also been getting angry over small issues that should have little effect on me. But lately, it's been having huge effects on me...It may sound like I'm a brat... \textbf{Thoughts}: I am confused about what is happening...I need an explanation for these changes to my mood...I feel out of control...I cannot control my behavior...my behavior and feelings are fused together...what I'm feeling is wrong... & Mental filter & The client is experiencing the complexity of emotional distress, which may cause the feature of "Mental filter" being overlooked.  \\
        \midrule\midrule
        \textbf{Case} & \textbf{\makecell[l]{Wrongly Recognized \\Disorder}} & \textbf{Reason} \\
        \midrule
        \textbf{Situation}: I have been with my girlfriend for 5 years and we have a history of a strong sexual intimacy connection. This has been getting worse over time. \textbf{Thoughts}: It is my girlfriend's sex drive that is the problem. & Personalization & This client attributes the issue solely to his girlfriend rather than examining internal or relational factors. The model's misidentification as “personalization” (taking excessive personal responsibility) could stem from misinterpreting the client’s statements. \\
        \midrule
        \textbf{Situation}: I am struggling at the moment and yet I am aware that I should be feeling good. \textbf{Thoughts}: What I am feeling and going through is unacceptable and wrong. There is a right and wrong way to go through this and I am doing it the wrong way. & All-or-nothing thinking & The core issue lies more in self-judgment and difficulty accepting emotional experiences, not binary thinking. The misidentification by the model could occur because both patterns involve rigid judgments. \\
        \midrule
        \textbf{Situation}: I am finding younger girls sexually arousing. \textbf{Thoughts}: I am worried about whether or not this will go away later in life, this is horrible, and I even feel guilty constantly, I cannot help it. & magnification & This patient is merely expressing reasonable concerns based on their current situation; it is rational and not an unjustified exaggeration. The misidentification by the model may caused by solely concentrating on the client's negative statements. \\
        \bottomrule
    \end{tabular}
    \caption{Error Cases of Cognitive Disorder Classification. (Llama-3.1-405B)}
\end{table*}
\endgroup

\begingroup
\small
\setlength{\tabcolsep}{4pt}
\begin{table*}[t]

    \centering
    \small
    \begin{tabular}{p{8cm}|l|p{4cm}}
        \toprule
        \textbf{Case} & \textbf{\makecell[l]{Unrecognized \\Core Belief}} & \textbf{Reason} \\
        \midrule
        \textbf{Situation}: She is everything I ever wanted in a woman and I am so happy to have her in my life. Unfortunately, I am not her first in many things, if anything at all, and that is very hurtful and distressing to me. \textbf{Thoughts}: If I'm not her first, she won't love me forever. She will leave me because I'm not special to her. She is everything to me. If she leaves me, I will be nothing. I won't ever be able to find someone as amazing as she is. I think she's lying to me about her virginity, which just means she is probably lying about her feelings for me, too, and it's only a matter of time before she realizes it and leaves me. & I am needy & The model may have concentrated on the client’s suspicions about the partner’s virginity and honesty, viewing the issue as insecurity or control, rather than emotional dependency.\\
        \midrule
        \textbf{Situation}: I met this guy a month ago, and we hung out and kissed, but in front of his sister, he told me I was too young for him and he only wanted to be friends. Now he supposedly has a girl that he spends all his time with after work and he doesn't text me anymore. \textbf{Thoughts}: I don't deserve love. There is something wrong with me. His new girl has something that I don't; that's why he is spending time with her and not me. I will be alone forever. I will always be rejected by everyone I care about. Nobody likes me. & I am needy & The model may have interpreted the client’s thoughts as purely low self-esteem or fear of rejection, rather than recognizing the underlying emotional need for constant validation.\\
        \midrule
        \textbf{Situation}: I have recently recovered from cancer, gained weight, and lack confidence in myself. I feel alone in my life. I still work but that is all there is. My wife and I get along but there is no emotional closeness between us. I have no close friends. \textbf{Thoughts}: I am a loser. I am a failure. Something is wrong with me. My wife deserves better than me. My wife doesn't love me anymore because I have gained weight. There is nothing enjoyable in my life, ever. There is absolutely nothing to look forward to. Me and my wife NEVER connect. I am all alone. Nobody loves me. This will never get any better. There is nothing I can do about it. The world is against me and it's out of my hands. I am a waste of space. Maybe cancer should have killed me. & I don’t deserve to live & The model might have missed the suicidal thoughts indicated by the statements like "There is nothing enjoyable in my life" and "Maybe cancer should have killed me". \\
        \midrule\midrule
        \textbf{Case} & \textbf{\makecell[l]{Wrongly Recognized \\Core Belief}} & \textbf{Reason} \\
        \midrule
        \textbf{Situation}: For the past 3 months I've been feeling really down, having mood swings, irritability -- I have no reason to be and I feel like I'm going crazy and that I can't talk to anybody about this because I'm being overly dramatic. \textbf{Thoughts}: Something is wrong with me. I am losing my mind. Nobody understands me. Everyone would reject me if they knew. & I am helpless & The client’s thought, “Nobody understands me,” suggests a need for connection and validation, which the model might have missed by emphasizing helplessness. \\
        \midrule
        \textbf{Situation}: When I go to the store, I believe that I hear people talking about me in their heads, as though I can hear their thoughts. \textbf{Thoughts}: I am capable of hearing other people's thoughts. People are talking to me in their heads. I am telepathic. I am certain of these things. Because I hear voices, they must be coming from other people. & I am out of control & The client’s thoughts reflect conviction (e.g., “I am telepathic”) rather than fear, indicating a belief in special abilities rather than being out of control. The model may not be very clear about the difference between the two. \\
        \midrule
        \textbf{Situation}: I have a problem of automatically looking at things like shiny objects or body parts. I don't have any bad intentions but people misunderstand me. \textbf{Thoughts}: This is something to be ashamed of. I always do it. Something is wrong with me. I can't stop looking at these things. People will reject me because of this habit. This habit is out of my control. & I am immoral & The client explicitly states they have no bad intentions, indicating the issue is about loss of control, not morality. The model may have neglected this information. \\
        \bottomrule
    \end{tabular}
    \caption{Error Cases of Fine-grained Core Belief Classification. (Llama-3.1-405B)}
\end{table*}
\endgroup

\clearpage
\section{Data Annotation Details}\label{sec:ann_details}
\begin{figure}[h]
\centering
\includegraphics[width=0.96\linewidth]{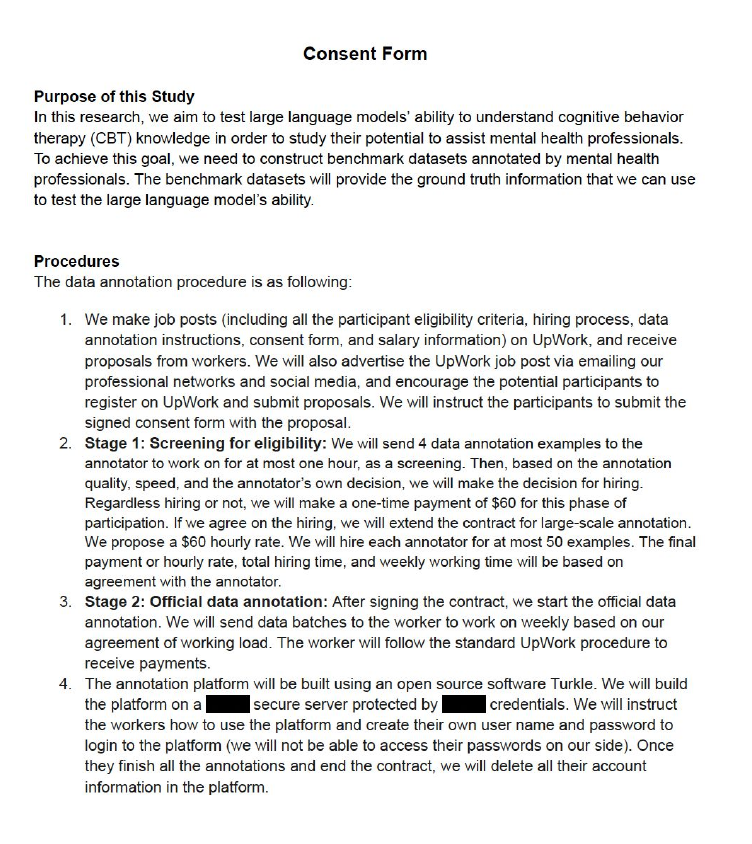}
\caption{Consent form page 1.}
\label{fig:ann_consent_1}
\end{figure}

\begin{figure}[h]
\centering
\includegraphics[width=0.96\linewidth]{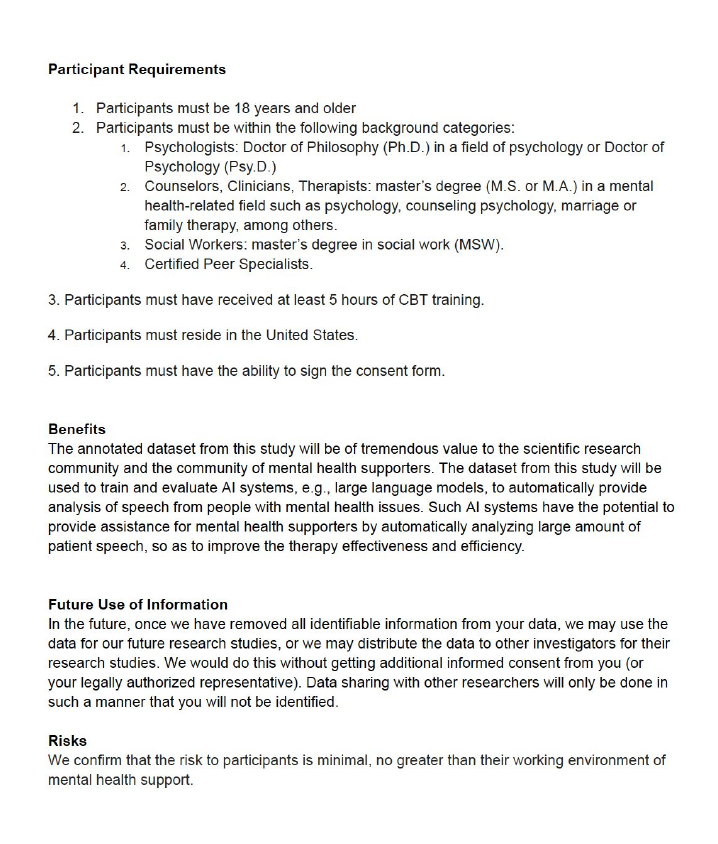}
\caption{Consent form page 2.}
\label{fig:ann_consent_2}
\end{figure}

\begin{figure}[h]
\centering
\includegraphics[width=0.96\linewidth]{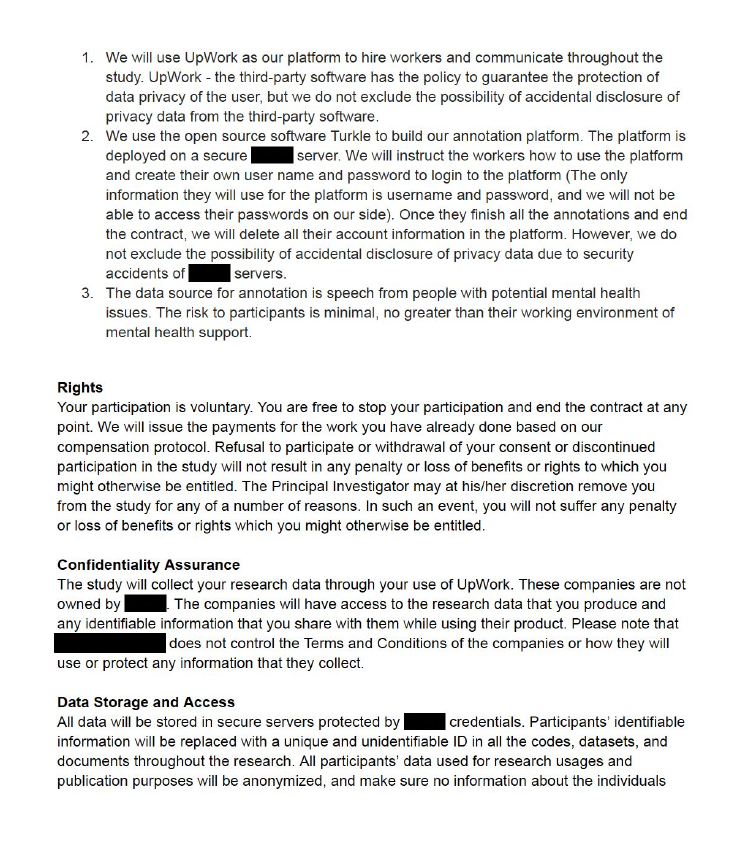}
\caption{Consent form page 3.}
\label{fig:ann_consent_3}
\end{figure}

\begin{figure}[h]
\centering
\includegraphics[width=0.96\linewidth]{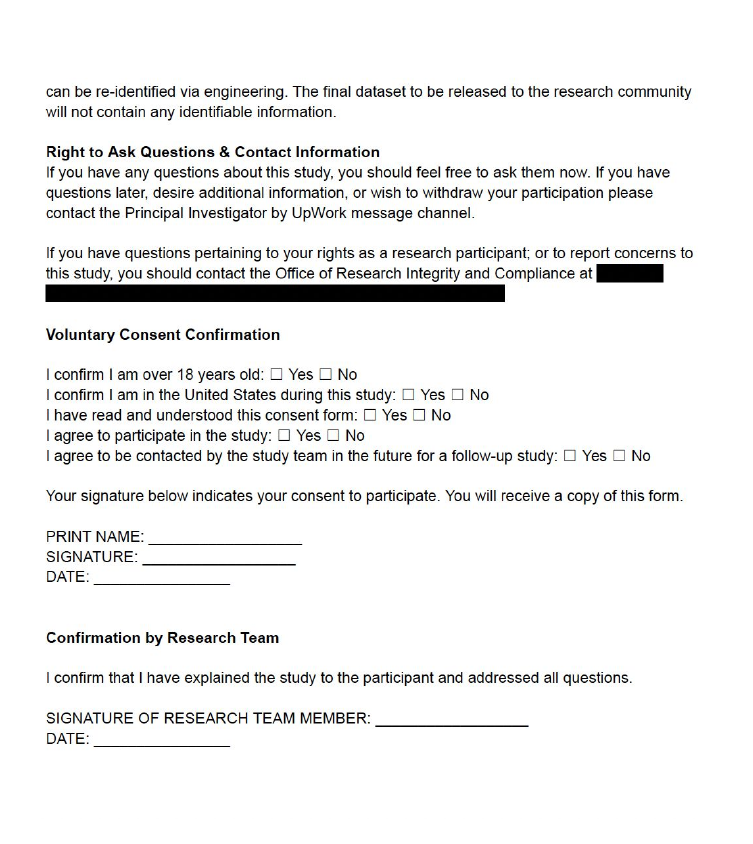}
\caption{Consent form page 4.}
\label{fig:ann_consent_4}
\end{figure}

\begin{figure*}[h]
\centering
\includegraphics[width=\linewidth]{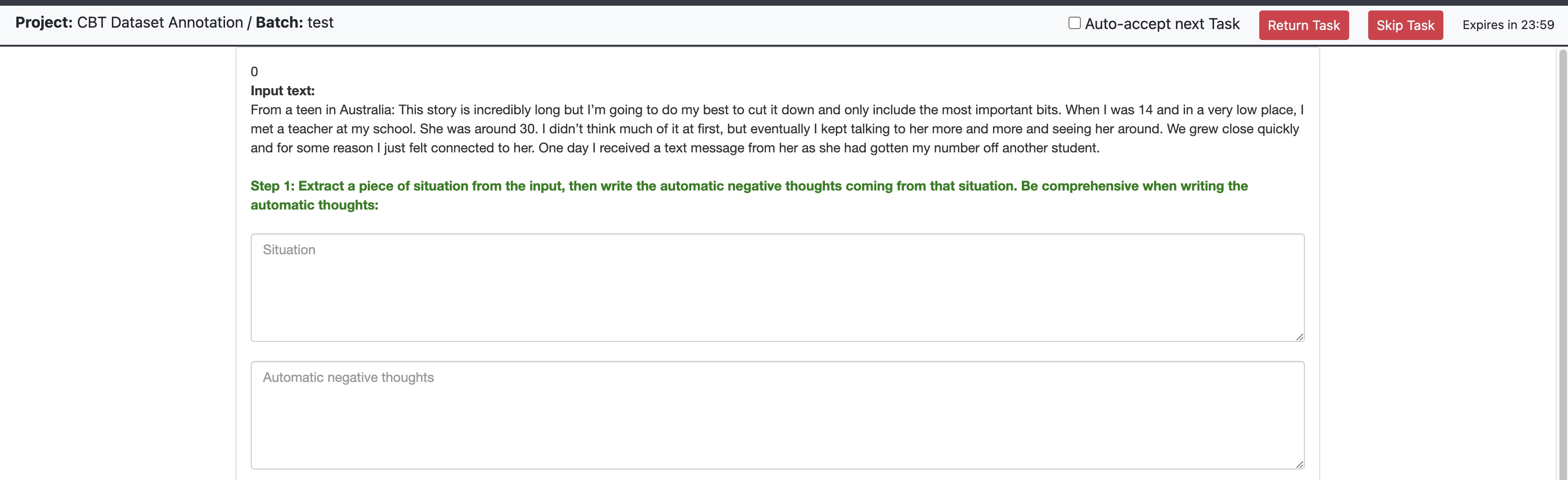}
\caption{Annotation interface of level \RN{2} tasks.}
\label{fig:ann_inter_1}
\end{figure*}

\begin{figure*}[h]
\centering
\includegraphics[width=\linewidth]{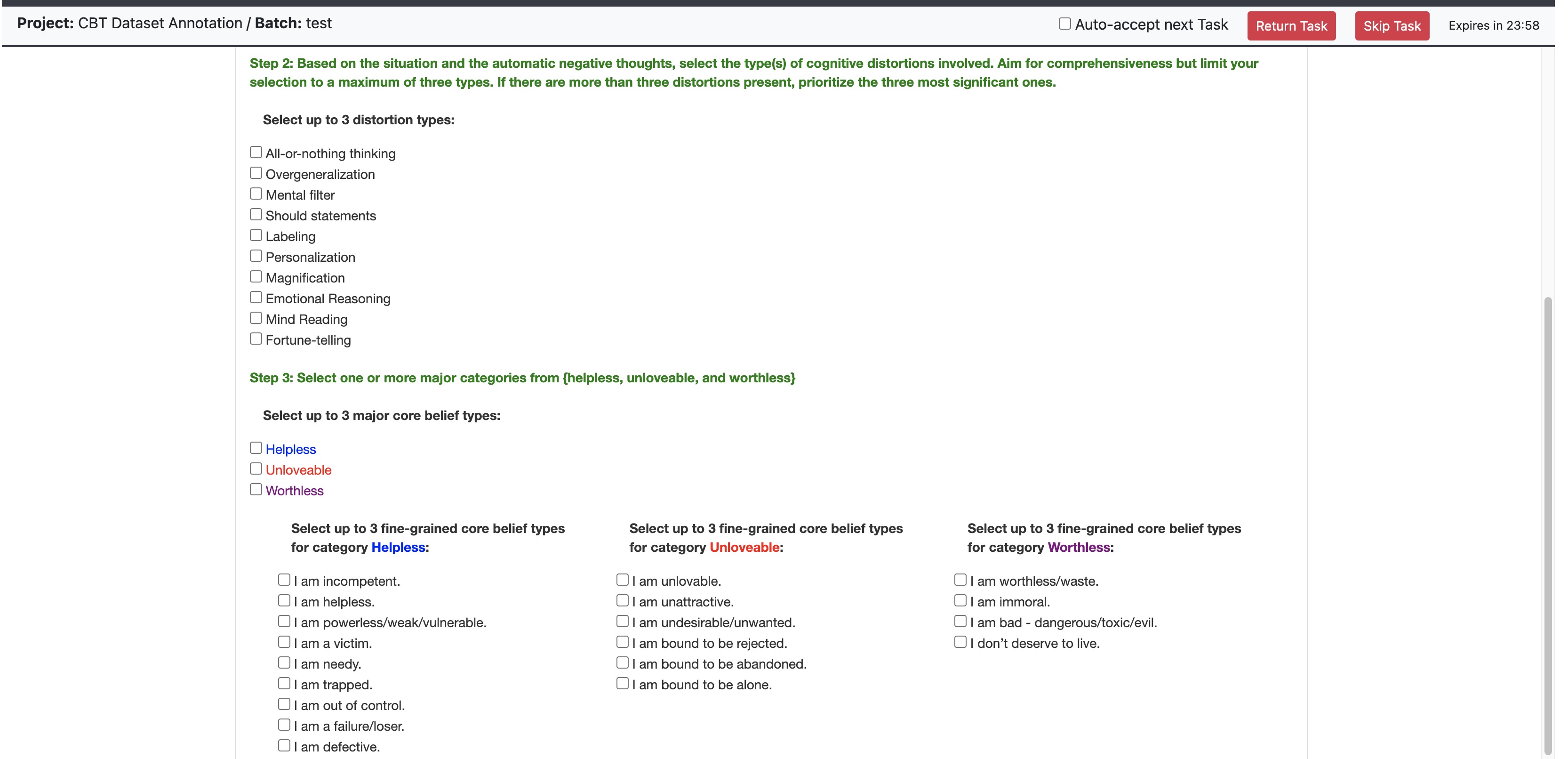}
\caption{Annotation interface of level \RN{2} tasks.}
\label{fig:ann_inter_2}
\end{figure*}

\begin{figure*}[h]
\centering
\includegraphics[width=\linewidth]{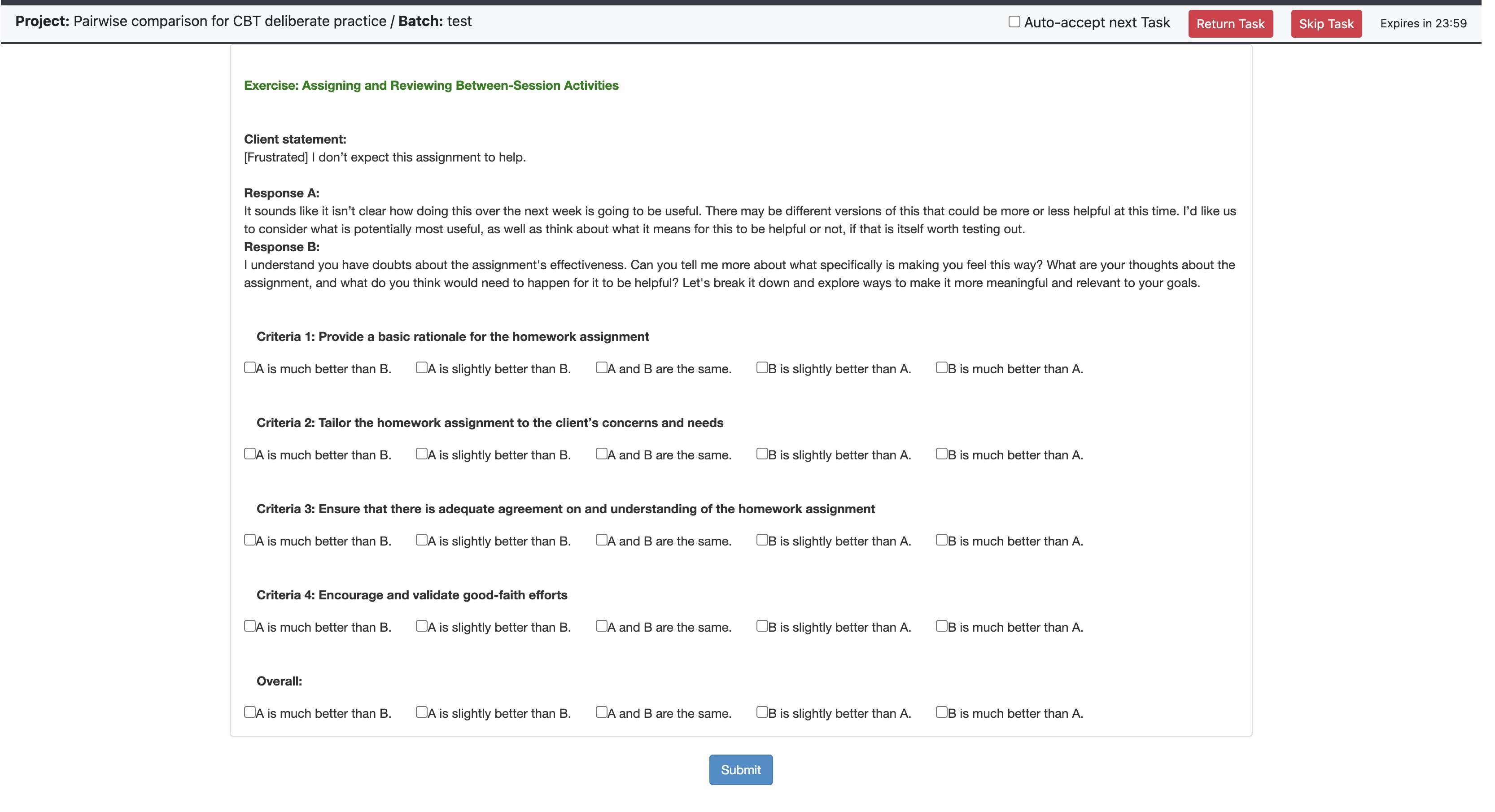}
\caption{Annotation interface of level \RN{3} task evaluations.}
\label{fig:ann_inter_3}
\end{figure*}

We use Turkle\footnote{https://github.com/hltcoe/turkle} to build our annotation interfaces. Figure~\ref{fig:ann_consent_1}, \ref{fig:ann_consent_2}, \ref{fig:ann_consent_3}, \ref{fig:ann_consent_4} shows our consent form presented to the annotators. Figure~\ref{fig:ann_inter_1} and \ref{fig:ann_inter_2} show our annotation interface for level \RN{1} and \RN{2} tasks. Figure~\ref{fig:ann_inter_3} shows our annotation interface for the evaluation of level \RN{3} task. 


\clearpage
\section{Level \RN{3} Case Study}
\begin{table*}[ht]
\centering
\resizebox{\textwidth}{!}{
\begin{tabular}{p{3cm}p{18cm}}
\toprule
\textbf{Statement} & You think I should set up some “little rewards” to start feeling less depressed. That seems kind of embarrassing and infantilizing. \\
\midrule
\textbf{GPT-4o} & I understand it might feel that way at first, but these small rewards are about acknowledging your accomplishments and creating positive reinforcement. They're designed to help you build momentum and foster a sense of achievement, which can gradually improve your mood and combat depression. Let's find rewards that feel meaningful and respectful to you. \\
\midrule
\textbf{Llama3.1-8B} & I understand your concern. In CBT, we often use small, achievable rewards to motivate and reinforce positive behaviors, but I can see how it might feel uncomfortable at first. Let's explore this further: what specifically makes you feel embarrassed or infantilized by the idea of setting up rewards? Is it the idea of receiving something tangible, or something else entirely? \\
\midrule
\textbf{Llama3.1-405B} & I understand your concern, but let's reframe the idea of rewards. It's not about treating yourself like a child, but about acknowledging and celebrating small achievements that can help build motivation and momentum. Think of it as a way to recognize your efforts and progress, rather than a childish reward system. Can we explore what kind of recognition or acknowledgment might feel more meaningful to you?\\

\bottomrule
\end{tabular}
}
\caption{Level  \RN{3} Case Study}\label{case-study-level-3}
\end{table*}

\clearpage
\section{Ethical Considerations}\label{sec:ethic}
\noindent\textbf{The Role of AI in Supporting Mental Health Professionals}:
\begin{itemize}
    \item Augmentation Over Replacement: We will emphasize that AI tools are designed to augment the capabilities of mental health professionals, not to replace them. This includes examples of providing therapists with data-driven insights into patients’ speech, prompting suggestions for therapy responses, etc.
    \item Training and Integration: Specialized training is necessary for mental health professionals to effectively integrate AI tools into their practice, ensuring they are equipped to use these technologies ethically and effectively.
\end{itemize}

\noindent\textbf{Safeguards Against Direct Interaction Without Supervision}:
\begin{itemize}
    \item Supervised Deployment Models: For application development regulations, AI-generated insights or interventions are always reviewed by a qualified professional before reaching a patient.
    \item Safety Protocols: Safety protocols must be designed to prevent AI systems from operating autonomously, including strict access controls, intervention thresholds, and mandatory oversight mechanisms.
\end{itemize}

\noindent\textbf{Transparency and Communications}:
\begin{itemize}
    \item Open Communication: Open channels of communication should be maintained with all stakeholders, including mental health professionals, patients, and regulatory bodies, ensuring transparency in how AI tools are developed and deployed.
    \item Explainability and Accountability: AI tools must provide clear reasoning for their decisions, especially in sensitive areas like mental health diagnosis or therapy recommendations. The lack of explainability can lead to mistrust or misuse.
    \item User Awareness: Patients and professionals interacting with AI tools must be informed that they are engaging with an AI system and understand its capabilities and limitations to prevent over-reliance or inappropriate application.
\end{itemize}

\noindent\textbf{Bias \& Fairness}:
\begin{itemize}
    \item Bias Mitagation: The output of AI tools should not be biased,  accounting for diverse linguistic, social, and cultural nuances while avoiding stereotypes or stigmatizing language that could harm or alienate individuals, particularly from underrepresented or marginalized groups.
    \item Accessibility: Ensuring that AI healthcare tools are accessible to underserved or marginalized populations, addressing disparities in healthcare availability.
\end{itemize}

\noindent\textbf{Safety \& Privacy}:
\begin{itemize}
    \item Error Mitigation: AI tools should minimize harm by reducing errors in medical advice or treatment recommendations. Misdiagnoses or inappropriate suggestions could cause significant harm to patients.
    \item Misinformation and Hallucination: AI tools in mental health care must prioritize accuracy and reliability by minimizing misinformation and hallucinations. These models should provide evidence-based, context-appropriate responses to avoid misleading users, as inaccurate information could harm individuals’ well-being or decision-making.
    \item Data Privacy and Confidentiality: AI tools in mental health care must uphold strict data privacy and confidentiality standards, ensuring that user interactions and sensitive information are securely stored, processed, and anonymized. Any data handling should comply with legal and ethical guidelines to protect users from breaches, misuse, or unauthorized access.
\end{itemize}

\end{document}